%% file: research_paper.tex
\documentclass{article}
\usepackage{iclr2025_conference,times}
\ClearShipoutPicture
\usepackage{graphicx}
\usepackage{booktabs}
\usepackage{multirow}
\usepackage{threeparttable}
\usepackage{tabularx}
\usepackage{algorithm}
\usepackage{algpseudocode}
\usepackage{amsmath,amssymb,bm}
\usepackage{xcolor}
\algrenewcommand\algorithmiccomment[1]{\{#1\}}
\newcommand{\Step}[1]{\State \textcolor{blue}{\textit{\{#1\}}}}
\AtBeginDocument{\thispagestyle{empty}}

\input{math_commands.tex}

\usepackage{hyperref}
\usepackage{url}

\title{Wavelet-Energy-Driven Dynamic Patching for Time Series Forecasting}

\author{Syed Hafiz Bin Salim Alsree \\
Department of Engineering Science \\
University of Oxford \\
\texttt{hafiz.alsree@eng.ox.ac.uk} \\
\And
Stefan Zohren \\
Department of Engineering Science \\
University of Oxford \\
\texttt{stefan.zohren@eng.ox.ac.uk}
\And
Stephen Roberts \\
Department of Engineering Science \\
University of Oxford \\
\texttt{sjrob@robots.ox.ac.uk}
}

\begin{document}

{\centering
{\LARGE\sc Wavelet-Energy-Driven Dynamic Patching for Time Series Forecasting\par}
\vskip 0.2in
{\bf Syed Hafiz Bin Salim Alsree}\\
Department of Engineering Science, University of Oxford\\
\texttt{hafiz.alsree@eng.ox.ac.uk}\\[0.6em]
{\bf Stephen Roberts}\\
Department of Engineering Science, University of Oxford\\
\texttt{stephen.roberts@eng.ox.ac.uk}\\[0.6em]
{\bf Stefan Zohren}\\
Department of Engineering Science, University of Oxford\\
\texttt{stefan.zohren@eng.ox.ac.uk}\\[0.6em]
\vskip 0.3in
}

\begin{abstract}
Patch-based transformers tokenise a time series into fixed-size, uniform windows, which commits the model to a single temporal resolution even though real signals are multi-scale. The discrete wavelet transform (DWT) is a natural alternative because it decomposes a signal across scales. A growing body of work makes the wavelet filter learnable, on the assumption that a more flexible wavelet forecasts better. That assumption has rarely been tested in isolation, because a single reported improvement usually credits three things at once. These are the use of any multi-resolution decomposition, the choice of wavelet and training the filter. We introduce a forecaster that turns a learnable-DWT per-scale energy map into variable-length patches by merging adjacent windows that share a dominant scale. We use it to run within-model ablations that vary one factor at a time. Three findings follow. First, most of the improvement is there before any filter is learned. A frozen wavelet, with no learning at all, beats a fixed-patch baseline on eight of the nine datasets by 1.5\% to 7.8\%, and on four of them it recovers all of the gain. Traffic is the one exception, where it is 1.2\% worse. Second, learning the filter gives a further improvement on only some datasets. It helps on three of the nine datasets and makes Traffic 2.7\% worse. Two of the three winners are finely sampled series whose high-pass band reshapes over the first few epochs and then settles, but hourly Electricity gains the most of all and none of the measures we examined accounts for it. Third, having variable-length patches helps, but exactly where their boundaries fall does not. A fixed-size uniform grid, even with its patch size tuned per task, is worse than our variable-length patches on 27 of 36 settings. Recomputing the boundaries from the learned filter at every step, rather than caching them once, never helps. The improvement in wavelet-based forecasting comes from the multi-resolution structure and the variable-length patches it induces, not from making the wavelet learnable. In practice, start from any valid wavelet, cache the patch boundaries once and treat training the filter as a per-dataset decision rather than a default.
\end{abstract}

\newpage
\section{Introduction}
\subsection{Motivation}

Long-term time-series forecasting supports planning in energy, climate, finance and healthcare, where decisions depend on predictions many steps beyond the next observation. Transformer-based long-term timeseries forecasting (LTSF) has converged on a common tokenisation step. Each channel-independent series is split into patches, the patches are embedded and a transformer models inter-patch dependencies \citep{nie2023patchtst}. PatchTST established that fixed-size patches with fixed stride are a strong and simple choice. Yet a single patch length encodes one fixed temporal resolution, while real signals are inherently multi-resolution. A slow trend and fast oscillation coexist within the same window and one patch length cannot match both. A patch long enough to span a slow trend blurs short, high-frequency bursts. One short enough to resolve those bursts fragments the trend across many tokens. Neither resolution is right for the whole series. Classical signal processing addresses this with multi-resolution analysis and the discrete wavelet transform (DWT) which separates a signal into approximation (low-frequency) and detail (high-frequency) components across scales. The DWT is therefore a natural basis for adaptive, resolution-aware tokenisation. This is the premise of our work. Our inspiration comes from language modeling, where single characters are too fine, whole words are too coarse and tokenisation settled on sub-word units in between. Recent work pushes past even that fixed vocabulary to tokenise directly from raw bytes. The Byte Latent Transformer \citep{pagnoni2024blt} forms dynamically-sized, content-driven patches, while autoregressive U-Nets \citep{videau2025bytes} pool raw bytes into a fixed multi-scale hierarchy of word and multi-word units. If dynamic tokenisation helps for text, an analogue built from the signal itself should help for timeseries. The same tokeniser could in turn help large language models read timeseries. 

\subsection{Our Approach}

To our knowledge, no prior forecasting method both (i) learns the wavelet filter and (ii) derives the patch boundaries from that learnable decomposition end-to-end. Ours does, and the same design lets us test the learnable-wavelet premise directly. Our approach rests on three design choices:

\begin{enumerate}
    \item using a multi-resolution wavelet decomposition at all.
    \item the choice of wavelet basis it starts from.
    \item training the filter versus leaving it fixed.
\end{enumerate}

Our model contains all three, so we can measure how much of its improvement comes from each one, the decomposition itself, the starting wavelet or the learning, rather than reporting a single combined number. 

The answer matters beyond our own model. Several recent methods spend parameters and training compute on learning the filter, and whether that investment pays off determines where future effort is best spent. This paper separates the three factors. We build a forecaster in which the wavelet's per-scale energy drives the patch boundaries directly, then use controlled ablations to ask where the improvement comes from. First, wavelet-driven patching beats fixed-size patches on eight of the nine benchmarks, and beats a tuned uniform grid built over the same decomposition on eight of the nine (§5.3). Second, most of that advantage comes from the decomposition and patching structure itself. The starting point must still be a valid wavelet, though which wavelet matters very little. Learning the filter end-to-end helps on some datasets, does nothing on others, and on one it hurts. These findings show where the benefit in wavelet-based forecasting comes from, which prior learnable-wavelet work has not separated across a benchmark panel. They are also consistent with recent evidence that a tuned fixed-patch baseline is hard to beat \citep{zucchi2026}.

\subsection{Contributions}
\begin{itemize}
    \item \textbf{Wavelet-energy-driven-variable-length patching.} We introduce a patching scheme driven by wavelet scale energy. A learnable DWT computes a per-scale energy map, adjacent windows that share a dominant scale are merged into variable-length patches (long over smooth stretches, short over detailed ones), and the patches feed a channel-independent transformer. No prior forecasting method both learns the wavelet filter and derives its patch boundaries from that filter.
    \item \textbf{Empirical validation across datasets and horizons.} The method matches or beats eight rerun baselines on most tasks, and is significantly better than six of the eight by a paired signed-rank test over the nine datasets (§4.2). It is level with the strongest baseline, and we report where it loses. Against four dynamic-patching methods rerun under the same three-seed protocol, it has the lowest MSE on three of seven datasets.
    \item \textbf{Disentangling decomposition structure, wavelet choice and filter learnability.} Prior learnable-wavelet work usually credits one improvement number to three choices at once. We separate them with controlled within-model ablations (§5.1, §5.2). The decomposition carries most of the gain. Learning the filter helps on three of the nine datasets and harms one. Recomputing the patch boundaries from the current filter is no better than caching them once (§5.1). The contribution lies in establishing which component the gain depends on, rather than in the architecture itself.
\end{itemize}

\subsection{Scope and framing}
This paper is both a method and an analysis. We introduce the patching mechanism (C1), evaluate it across datasets and forecast horizons (C2) and use controlled within-model ablations, varying one factor at a time under a fixed architecture to locate where the benefit comes from (C3). Because each comparison varies a single factor within one architecture, our conclusions about where the gain originates do not depend on how well any external baseline was tuned. The remainder of the paper reviews related work (§2), details the method (§3), reports the main results (§4), and presents the controlled analysis and ablations (§5).

Code, configurations and the scripts for every baseline run are available at \url{https://github.com/hafizalsree/dwt-dynamic-patching}.

\section{Background and Related Work}

\paragraph{Wavelet background.} A single-level DWT applies a low-pass filter $h$ and a high-pass filter $g$, each followed by downsampling-by-two, splitting a length $n$ into approximation $a$ and detail $d$ coefficients of length $n/2$. Iterating the filter bank on the approximation branch yields a multi-level decomposition. For an orthogonal wavelet the high-pass filter is the quadrature mirror of the low-pass filter ($g[k] = (-1)^{(k)}\cdot h[N-1-k]$), so a classical wavelet is specified by $h$ alone. Section~\ref{sec:learnable} explains why we do not impose that tie. Two properties define a wavelet family. \textbf{Vanishing moments (VMs)} count the polynomial degrees the detail filter $g$ annihilates. \textbf{Filter width (FW)} trades time-domain against frequency-domain localisation. Daubechies wavelets maximise VMs for a given support. Higher $N$ gives a smoother basis and wider support.

\paragraph{Wavelet/Frequency decomposition with a fixed transform.} These methods add a spectral decomposition but keep the transform fixed and learn only the downstream mapping. Frequency-domain transformers decompose with a fixed Fourier basis \citep{zhou2022fedformer}. Recent wavelet methods keep a fixed wavelet and learn a map in the wavelet domain \citep{xie2025swift}, a subband weighting router \citep{wang2025wavetuner} or a patch and mix method \citep{murad2025wpmixer}. None of these methods learns the filter itself. 

\paragraph{Learnable wavelets.} A smaller line of work makes the wavelet itself trainable. mWDN \citep{wang2018mwdn} turns multi-level DWT filter coefficients into trainable parameters and AdaWaveNet \citep{yu2024adawavenet} replaces the fixed wavelet with a trainable lifting scheme for forecasting. The most recent, Sonnet \citep{shu2026sonnet} makes the wavelet transform itself learnable and applies it to the input representation, pairing it with a spectral (Koopman) operator for multivariate forecasting, though it learns the transform on embeddings rather than using it to place patch boundaries. The idea of learning the wavelet filter itself is not unique to forecasting. In signal processing, the fully learnable deep wavelet transform \citep{michau2022fldwt} makes the filter coefficients trainable end-to-end under a conjugate-quadrature-filter constraint and WavSpa \citep{zhuang2024wavspa} makes the wavelet learnable three ways, through direct, orthogonal and lifting scheme parameterisation inside an attention model. The premise behind all of them is intuitive. A learnable wavelet is more flexible so it should forecast better. But as §5 shows, it has rarely been tested in isolation to our knowledge, not across a full benchmark panel.

\paragraph{Dynamic Patching.} Several methods move past fixed patches by adapting where patches fall or how large the patches are. Pathformer \citep{chen2024pathformer} routes among patch sizes per input, EntroPE places boundaries at predictive entropy transitions \citep{abeywickrama2025entrope}, TimeMosaic \citep{ding2025timemosaic} varies patch granularity by local information density, while HDMixer \citep{huang2024hdmixer} learns deformable length-extendable patch boundaries. Reported gains are mixed, however. A recent study finds no consistent advantage for adaptive patching over a tuned fixed-patch baseline \citep{zucchi2026}.

\paragraph{Positioning.} This analysis \citep{zucchi2026} covers only methods that place their cuts in the time domain from the raw signal. None of the methods it tests derives patch boundaries from a frequency decomposition, which is what ours does, so its conclusion does not transfer directly to our setting. We therefore run its within-model test on our own model in §5.3, and test its count-versus-position question directly by replacing our cuts with random ones at the same count. We did not find such a control in the dynamic patching methods we compare with in Table~\ref{tab:dynpatch}.

\section{Methodology}
\begin{figure*}[hb!]
  \centering
  \includegraphics[width=\textwidth]{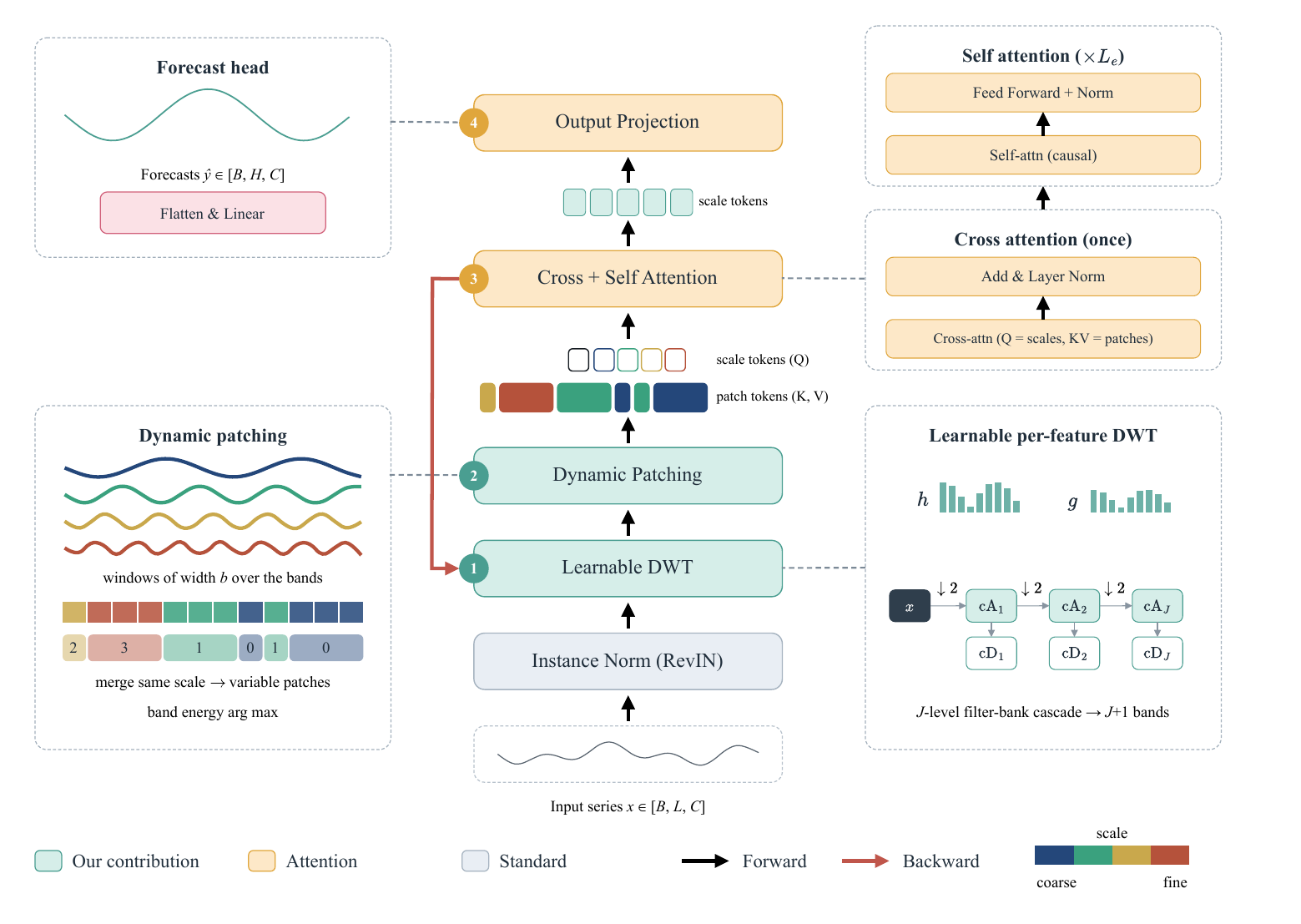}
  \caption{End-to-end architecture. The learnable DWT decomposes each channel, per-scale energy sets variable-length patch boundaries, and the merged patches feed a transformer backbone with a scale embedding.}
  \label{fig:architecture}
\end{figure*}

We forecast a multivariate series by replacing the fixed, uniform patches of PatchTST with a data-driven, multi-resolution tokenisation. Figure~\ref{fig:architecture} shows the pipeline and this section derives it end-to-end. We first define the wavelet filters and the decomposition, then per-scale reconstructions and the energy-driven patch boundaries they induce, then the tokens, the attention backbone, the forecast and training objective.

\textbf{Notation.} The input is $X\in \mathbb{R}^{C\times L}$, with $C$ channels and $L$ timesteps, and we forecast the next $H$ steps of every channel. Each channel is processed separately with shared weights, so we describe the model for one channel $x\in \mathbb{R}^{L}$. The model width is $d$.

\subsection{Wavelets, scaling functions and filter banks}
A multi-resolution analysis represents a signal at a hierarchy of resolutions. Two functions generate it. A scaling function $\varphi(t)$ which captures coarse, low frequency content, and a wavelet function $\psi(t)$ which captures fine, high frequency detail. These satisfy the two-scale dilation equations, built from dilated and shift copies of the scaling function $\varphi$:

\begin{equation}
    \varphi(t)=\sqrt2\sum_{k} h[k]\,\varphi(2t-k),\qquad \psi(t)=\sqrt2\sum_{k} g[k]\,\varphi(2t-k)
\end{equation}

The weights are two discrete filters of length $N$, a low-pass filter $h\in\mathbb{R}^{N}$ and a high-pass filter  $g\in\mathbb{R}^{N}$. The filters therefore fully determine the basis, so choosing ($h$,$g$) chooses the wavelet. In a classical orthogonal wavelet the two are tied by the quadrature-mirror-filter (QMF) relation, $g[k]=(-1)^{k}h[N-1-k]$, which, together with the usual orthonormality conditions on $h$, guarantees orthogonality and perfect reconstruction.

On a discrete signal the transform is realised as a filter bank. Let $a_{0}=x$ be the input. One decomposition level convolves $a_{0}$ with each filter and then keeps every other sample (downsampling by two):

\begin{equation}
    a_{1}[n]=\sum_{k} h[k]\,a_0[2n-k],\qquad d_{1}[n]=\sum_{k} g[k]\,a_0[2n-k]
\end{equation}

This produces an approximation $a_{1}$ (the low-frequency part) and a detail $d_{1}$ (the high-frequency part), each of length about $L/2$. To obtain several resolutions we recurse on the approximation branch only, $a_{1}\to (a_{2},d_{2})$, then $a_{2}\to (a_{3},d_{3})$ and so on. After $J$ levels the signal is represented by one approximation and $J$ detail bands, 

\begin{equation}
    \mathrm{DWT}_J(x)=\{\,a_J,\;d_J,\;d_{J-1},\;\dots,\;d_1\,\},
\end{equation}

ordered from coarsest ($a_{J},d_{J}$) to finest ($d_{1}$). We write the number of bands as $S=J+1$.

\subsection{Learnable, unconstrained filters}\label{sec:learnable}
Rather than fix ($h,g$) to an off-the-shelf wavelet, we treat them as trainable parameters initialised to a named wavelet (db8) and updated by back-propagation alongside the rest of the model. Two choices matter. First, we drop the QMF tie. Once the filters are trained they are no longer a true wavelet, only a wavelet-initialised filter bank. We still call it a DWT for readability. We learn $h$ and $g$ independently and impose no orthogonality or perfect-reconstruction constraint because the transform is used only to produce features for the forecaster. Invertibility is unnecessary and dropping the constraints improves accuracy on three of the four ETT datasets we tested  (Appendix~\ref{app:qmf}). Second, we use per-feature filters. Each channel $c\in \{1,...,C\}$ has its own pair ($h^{(c)}, g^{(c)}$) so different variables may adopt different decompositions. Convolutions are evaluated in the frequency domain (an FFT backend),

\begin{equation}
    a_{j+1}=\downarrow_2\!\Big(\mathcal F^{-1}\big(\mathcal F(a_j)\odot \mathcal F(h)\big)\Big),
\end{equation}

where $\mathcal F$ is the FFT, $\odot $ the point-wise product and $\downarrow_2$ downsampling by two. Finally, we define three modes that serve as the control analysed in §5:

\begin{enumerate}
    \item Fixed: $(h, g)$ frozen at their wavelet initialisation. The decomposition is used, but nothing about the filter is learned.
    \item Learned: $(h, g)$ learned, with the patch boundaries computed from the initialisation filters and then held fixed.
    \item Dynamic: $(h, g)$ learned, with the boundaries recomputed from the current filters at every forward pass.
\end{enumerate}

Our main model is the \textbf{Learned} mode. Dynamic is analysed as an ablation in §5.1.

\subsection{Per-scale reconstruction}
The bands above have different lengths ($d_{1}$ has length $\approx L/2, d_{2}\approx L/4$, and so on), which makes them hard to compare on the time axis. We therefore map each band back to the original length $L$. We write the multi-level transform as $(c_{\mathrm a}, c_{0}, c_{1},\dots, c_{J-1})=\mathrm{DWT}(x)$, with $c_{\mathrm a}$ the approximation and $c_{0},\dots,c_{J-1}$ the detail bands ordered coarsest to finest. We isolate detail band $s$ by setting every other band to zero and performing the inverse DWT on the result:

\begin{equation}
    r_s=\mathrm{IDWT}\big(0,\dots,0,\;c_s,\;0,\dots,0\big)\in\mathbb R^{L},\qquad s=0,\dots,J-1
\end{equation}

Inverse DWT (IDWT) is the synthesis filter bank of §3.1 run with the same learned filters. Before training it inverts the analysis bank exactly, so $r_s$ is the part of the signal in band $s$. After training it is no longer an exact inverse, so we treat $r_s$ as a learned per-band feature at the original length, not as a true decomposition of $x$. We index the detail bands from $s=0$ (coarsest) to $s=J-1$ (finest). The approximation band, the overall trend, feeds the scale tokens (§3.5) but is not used for the patch boundaries since it has no localised, time-specific structure.

\subsection{Scale-energy-driven boundaries and soft scale weights}

We now turn the per-scale reconstructions into patch boundaries. Divide the time axis into contiguous base windows of fixed width $b$ (a hyper-parameter, e.g. $b=8$), so window $w$ covers indices [$wb, (w+1)b$] for $w=0,...[L/b]-1$.

\textbf{Hard assignment used in the forward pass.} For each window we measure how much energy each detail band carries inside it and label the window with the band that dominates:

\begin{equation}
    E_w[s]=\sum_{t=wb}^{(w+1)b-1} r_s(t)^2,\qquad m_w=\arg\max_{s\in\{0,\dots,J-1\}} E_w[s]
\end{equation}

\begin{figure}[t]
  \centering
  \includegraphics[width=\linewidth]{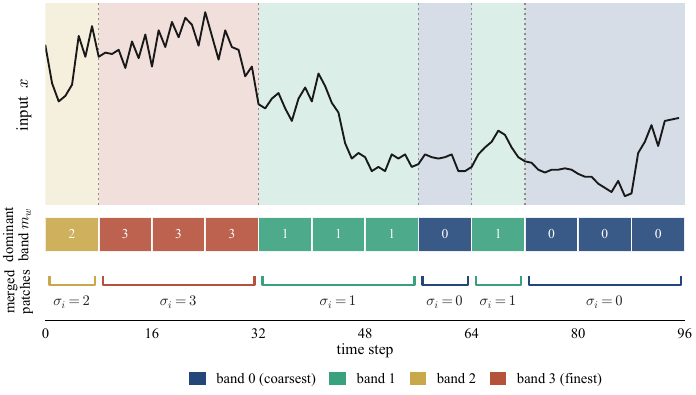}
  \caption{\textbf{Scale-energy-driven boundary patching (hard assignment).} A real input window (ETTm1, first channel, $L{=}96$, base-window width $b{=}8$, $J{=}4$ detail bands). \emph{Top:} the input $x$, shaded by the patch each region is assigned to. \emph{Middle:} the dominant detail band $m_w$ of each base window (the $\arg\max$ of per-band energy $E_w[s]$). \emph{Bottom:} adjacent windows sharing a dominant band merge into variable-length patches. Patch $i$ has band tag $\sigma_i$ and length $L_i$ (its width). A band that stays dominant over several windows forms a longer patch. Rapidly changing dominance yields shorter patches.}
  \label{fig:patching}
\end{figure}

We then group the windows into patches by scanning left to right and merging neighbouring windows that share the same dominant band, starting a new patch whenever the dominant band changes (Figure~\ref{fig:patching}). A run of windows all dominated by the same band becomes a single patch. This yields $M\le M_{max}$ patches, where $M_{\max}=L/b$ is the number of base windows. It is the maximum number of patches the merge can produce, so it is a padding length rather than a constraint. Sequences shorter than $M_{\max}$ are zero-padded and masked in the attention. Patch $i$ occupies a contiguous range $P_i$ of the time axis of length $L_i$ (its number of merged windows times $b$) and carries the band tag $\sigma_i$, the dominant band shared by those windows. Patch length therefore adapts to the signal. Where one band stays dominant across many windows, those windows merge into a long patch. Where the dominant band changes from window to window, they stay split into short patches. On the nine benchmarks merging uses 48\% to 63\% fewer tokens than the uniform grid it starts from.

\textbf{Soft assignment used only to pass gradient.} The $\textbf{arg max}$ above is not differentiable. In parallel we therefore compute a soft version, a softmax over bands within each window, at temperature $\tau$:

\begin{equation}
    \tilde w_w[s]=\frac{\exp\!\big(E_w[s]/\tau\big)}{\sum_{s'}\exp\!\big(E_w[s']/\tau\big)}
\end{equation}

Unlike $m_w$, the weights $\tilde{w}_w$ are a smooth function of the band energies $E_w$ and hence of the filters ($h, g$) through $r_s$. As $\tau\to 0$ the softmax concentrates on the dominant band (recovering the hard label). Larger $\tau$ spreads weight across bands. These weights do not change the forward boundaries.

\subsection{Tokenisation with a straight-through scale embedding}
Two sets of tokens enter the Transformer, patch tokens and scale tokens.

\begin{figure}[t]
  \centering
  \includegraphics[width=\linewidth]{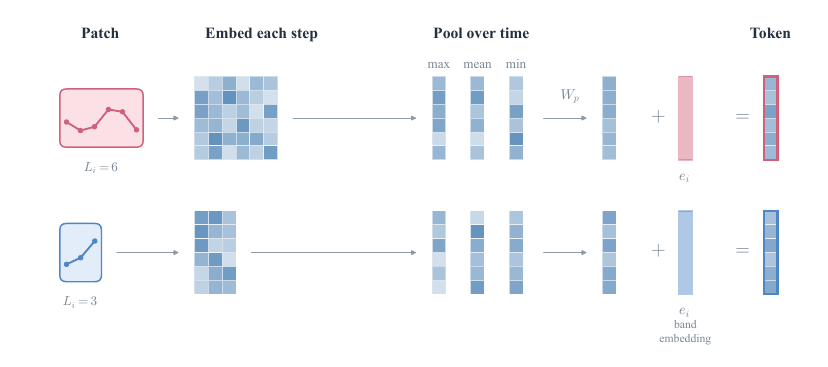}
  \caption{\textbf{Patch-token construction.} The variable-length patches from §3.4 (lengths $L_i$ differ) are each turned into a fixed-size token. We embed every time step to width $d$, then pool the patch's steps with the element-wise max, mean, and min \emph{over time} (collapsing the variable length $L_i$), project to $d$ with $W_p$, and add the learned band embedding $e_i=\mathrm{ScaleEmb}[\sigma_i]$ for the patch's scale $\sigma_i$. Every patch, of any length $L_i$ becomes a token $p_i\in\mathbb{R}^d$ of the same size.}
  \label{fig:tokens}
\end{figure}

\textbf{Patch tokens.} Patch $i$ contains the variable-length slice $x_{P_i}\in\mathbb{R}^{L_i}$ and the lengths $L_i$ differ from patch to patch. The Transformer, however, requires every token to be a vector of the same fixed width $d$. Each slice would present a different shape, so the slices cannot be consumed directly. We therefore map every patch to a fixed-size descriptor by pooling over its timesteps (Figure~\ref{fig:tokens}). Each step is first lifted to width $d$ by a shared map $W_{in}:\mathbb{R}\to\mathbb{R}^d$. We then summarise the patch by the element-wise maximum, mean and minimum of these vectors across its $L_i$ steps, concatenate the three, project with $W_p$ and add a band embedding $e_i$:

\begin{equation}
    v_t=W_{\mathrm{in}}\,x_t,\qquad
p_i=W_p\big[\;\max_{t\in P_i} v_t\;;\;\operatorname{mean}_{t\in P_i} v_t\;;\;\min_{t\in P_i} v_t\;\big]+e_i\in\mathbb R^{d}
\end{equation}

Pooling along the time axis makes the representation length-invariant, so patches of any $L_i$ yield tokens of the same width $d$. The three statistics are complementary. The mean captures the patch's overall level and maximum and minimum its range, so the descriptor preserves the patch's main characteristics regardless of how many steps it contains.

This descriptor is computed from the signal values alone and so does not record which frequency band the patch belongs to. We therefore also add a band embedding $e_i=\mathrm{ScaleEmb}[\sigma_i]$, a learned vector drawn from a table $\mathrm{ScaleEmb}\in\mathbb{R}^{S\times d}$ with one entry per band and indexed by the patch's scale tag $\sigma_i$ so that each token records which scale it came from alongside its pooled statistics.

\textbf{Scale tokens.} The $J$ detail reconstructions $r_0,...,r_{J-1}$ together with the approximation $r_a$ are each mapped to width $d$ by a learned projection $W_s$, giving $S=J+1$ scale tokens $u_s=W_sr_s$ that summarise the signal at each resolution.

The straight-through band embedding $e_i$ in the patch token is where the filter's gradient enters the patching. Let $\text{ScaleEmb}\in\mathbb{R^{S\times \text{d}}}$ be a learned embedding table, one row per band. Let $W_i$ be the set of base windows merged into patch $i$, so $|W_i|=L_i/b$. We define a hard embedding (the row of the patch's dominant band) and a soft embedding (the soft weights averaged over the windows of the patch, mixing the detail-band rows),

\begin{equation}
    e^{\text{hard}}_i=\mathrm{ScaleEmb}[\sigma_i],\qquad
e^{\text{soft}}_i=\Big(\tfrac1{|W_i|}\!\!\sum_{w\in W_i}\!\tilde w_w\Big)\,\mathrm{ScaleEmb}[0{:}J],
\end{equation}

and combine them with a straight-through estimator controlled by a strength $\alpha\ge 0$:

\begin{equation}
    e_i=e^{\text{hard}}_i+\alpha\big(e^{\text{soft}}_i-\operatorname{stopgrad}(e^{\text{soft}}_i)\big)
\end{equation}

In the forward pass $e_i = e^{\text{hard}}_i$ exactly, so the token carries the hard scale tag. In the backward pass the stopgrad leaves only the soft branch, scaled by $\alpha$, so gradient flows through $e_i^{soft}\to \tilde{w}_w \to E_w \to r_s \to (h, g)$. Hence $\alpha=0$ recovers the pure arg max model and the filter gets no gradient through the patching. With $\alpha > 0$ we get the dynamic-patching gradient update, and $\tau$ controls how sharp the soft signal is. This path is active only in the Dynamic mode. In the Fixed and Learned modes the boundaries are cached, so $\alpha$ has no effect.

\subsection{Cross- and Self-attention}
The patch tokens $P=[p_1,...,p_{M_{max}}]$ and the scale tokens $U=[u_0,...,u_J]$ are combined by a cross-attention layer. The scale tokens act as queries and the patch tokens as keys and values, so each resolution gathers the patch evidence relevant to it:

\begin{equation}
    C=\operatorname{CrossAttn}\big(Q=U,\;K=V=P\big)\in\mathbb R^{S\times d}
\end{equation}

A transformer encoder with causally masked self-attention then mixes these $S$ tokens, $Z=\text{SelfAttn}(C)$.

\newpage
\subsection{Forecast head, objective and gradient}
The encoder output $Z$ is flattened and mapped to the $H$-step forecast by a learned linear head, then de-normalised by the RevIN layer \citep{kim2022revin}. All parameters, meaning the learnable filters (in the learned modes), the token projections and embeddings, the attention weights and the head are trained jointly to minimise the mean-squared forecast error over the horizon:

\begin{equation}
    \mathcal L=\frac1H\big\lVert\hat y-y\big\rVert_2^2=\frac1H\sum_{t=1}^{H}\big(\hat y_t-y_t\big)^2
\end{equation}

Algorithm~\ref{alg:wedp} summarises the full pipeline. Appendix~\ref{app:detail} gives the step-by-step version.

\begin{algorithm}[h!]
\caption{Wavelet-energy-driven dynamic patching. Sub-routines (\textsc{small caps}) are derived in the cited sections.}
\label{alg:wedp}
\begin{algorithmic}[1]
\Require Series $X\in\mathbb{R}^{C\times L}$, target $Y$; per-channel learnable filters $\{h^{(c)},g^{(c)}\}$ (init to the task's tuned wavelet, db8 by default; QMF dropped); levels $J$ ($S{=}J{+}1$ bands); base-window width $b$; model width $d$; horizon $H$.
\Statex
\While{not converged}
  \State Sample $(X,Y)$; process each channel $x\in\mathbb{R}^{L}$ independently; \ $x\gets\mathrm{RevIN}(x)$
  \Step{Step 1: Multi-resolution decomposition (\S3.2--3.3)}
  \State $\{r_{\mathrm a},r_0,\dots,r_{J-1}\}\gets \textsc{DWT-Reconstruct}_{h,g}(x)$ \Comment{$S$ bands, each rebuilt at length $L$ (\S3.3)}
  \Step{Step 2: Scale-energy dynamic patching (\S3.4)}
  \State $\{m_w,\tilde w_w\}\gets \textsc{ScaleEnergy}(r_0,\dots,r_{J-1},b)$ \Comment{dominant band and soft weights per window; every step in Dynamic, once and cached in Fixed and Learned (\S3.4)}
  \State $\{P_i,\sigma_i\}_{i=1}^{M}\gets \textsc{MergeRuns}(m)$ \Comment{same-band windows $\to$ variable-length patches (\S3.4, Fig.~\ref{fig:patching})}
  \Step{Step 3: Tokenisation (\S3.5)}
  \State $p_i\gets \textsc{PatchToken}(x,P_i,\sigma_i,\tilde w)$ \Comment{pool + position embedding + straight-through band embedding (\S3.5)}
  \State $u_s\gets W_s\,r_s$ \Comment{one scale token per band (\S3.5)}
  \Step{Step 4: Attention and forecast (\S3.6--3.7)}
  \State $Z\gets \mathrm{SelfAttn}\big(\mathrm{CrossAttn}(Q{=}U,\,K{=}V{=}P)\big)$ \Comment{scales attend to patches, then causal self-attention over scales (\S3.6)}
  \State $\hat Y\gets \mathrm{RevIN}^{-1}(\mathrm{Head}(Z))$;\quad $\mathcal L\gets\tfrac1H\lVert\hat Y-Y\rVert_2^2$ \Comment{(\S3.7)}
  \Step{Step 5: Update}
  \State Update $W_{\mathrm{in}},W_p,W_s,\mathrm{ScaleEmb}$, encoder, head via $\nabla\mathcal L$, and $\{h^{(c)},g^{(c)}\}$ in Learned and Dynamic
\EndWhile
\end{algorithmic}
\end{algorithm}

\newpage
\section{Experiments}
\label{results}

\subsection{Setup}

\textbf{Datasets \& splits.} We evaluate nine standard multivariate benchmarks, ETTh1/h2/m1/m2, Weather, Exchange, ILI, Electricity (ECL, 321 channels) and Traffic (862 channels) at a single common lookback ($L=96$, $L=36$ for ILI) and four horizons (\{96, 192, 336, 720\}, and \{24, 36, 48, 60\} for ILI). We follow the standard chronological Time-Series-Library \citep{wu2023timesnet} splits, 12/4/4 months train/validation/test for ETT and 70/10/20 for the other datasets. Exact split sizes and per-dataset statistics are in the Appendix~\ref{app:datasets}.

\textbf{Baselines.} We compare against eight models from the transformer, MLP, linear and convolutional families, PatchTST \citep{nie2023patchtst}, iTransformer \citep{liu2024itransformer}, TimeMixer \citep{wang2024timemixer}, TimeXer \citep{wang2024timexer}, TimesNet \citep{wu2023timesnet}, DLinear \citep{zeng2023dlinear}, Autoformer \citep{wu2021autoformer} and the vanilla Transformer \citep{vaswani2017transformer}. To avoid the mixed-lookback/copied-number confound, we re-run every baseline ourselves in one harness at $L{=}96$. Where the authors publish a configuration at that lookback we run it unchanged, which covers 60 of the 72 baseline cells, and the remaining 12 transfer each model's nearest published configuration (Appendix~\ref{app:provenance}). We further compare against four dynamic-patching methods, EntroPE~\citep{abeywickrama2025entrope}, Pathformer~\citep{chen2024pathformer}, TimeMosaic~\citep{ding2025timemosaic} and HDMixer~\citep{huang2024hdmixer}, the closest prior work to ours. Each is run from its released code with its published configuration, unchanged except for the random seed, and these four methods and ours are reported as the mean over the same three seeds \{2021, 42, 123\}. Where a released script uses a longer lookback (TimeMosaic 320, HDMixer 336) we set it to 96 and keep all other settings. Appendix~\ref{app:provenance} records the configuration source of every cell.

\textbf{Our model.} Our main model is the \textbf{Learned} mode (§3), a per-feature learnable DWT with the QMF tie dropped. The Fixed, Learned and Dynamic modes form the ablation sequence in §5.

\textbf{Training \& optimisation.} Our models minimise MSE loss with the Adam optimiser (initial learning rate $10^{-4}$; $5\times10^{-4}$ for ILI) under the \texttt{type1} schedule. Batch size is 32 except on the two many-channel datasets, where GPU memory limits it to 16 (ECL) and 4 (Traffic). We train for up to 10 epochs with early stopping (patience 3) on the validation loss and restore the best-validation checkpoint for testing. Every other model uses the optimiser, batch size and training budget of its own published configuration. Results for our model on all nine datasets are averaged over three seeds \{2021, 42, 123\}. Full training and implementation details are in Appendix~\ref{app:impl}.

\textbf{Evaluation.} We report MSE and MAE, averaged over the four horizons in Table~\ref{tab:main} and given per-horizon in Appendix~\ref{app:perhorizon}. Significance is assessed by a paired Wilcoxon signed-rank test on the nine per-dataset mean MSEs at $p<0.1$. We do not test over the 36 (dataset $\times$ horizon) tasks, because the four horizons of a dataset share its training data and configuration and are not independent.

\subsection{Main results}

\begin{table}[h!]\centering\small\setlength{\tabcolsep}{3pt}
\caption{Main results: MSE / MAE averaged over horizons, lookback $L{=}96$ (ILI $L{=}36$). Best in \textbf{bold}, second best \underline{underlined}. Ours is the Learned variant (§3.2), the per-task mean over seeds \{2021, 42, 123\}. Per-dataset standard deviations are in Appendix~\ref{app:datasets}.}
\label{tab:main}
\resizebox{\textwidth}{!}{%
\begin{tabular}{lcccccccccccccccccc}
\toprule
Dataset & \multicolumn{2}{c}{Ours} & \multicolumn{2}{c}{Autoformer} & \multicolumn{2}{c}{PatchTST} & \multicolumn{2}{c}{TimesNet} & \multicolumn{2}{c}{Transformer} & \multicolumn{2}{c}{TimeMixer} & \multicolumn{2}{c}{TimeXer} & \multicolumn{2}{c}{iTransformer} & \multicolumn{2}{c}{DLinear} \\
 & MSE & MAE & MSE & MAE & MSE & MAE & MSE & MAE & MSE & MAE & MSE & MAE & MSE & MAE & MSE & MAE & MSE & MAE \\ \midrule
ETTh1 & \textbf{0.421} & \textbf{0.438} & 0.515 & 0.495 & 0.449 & 0.449 & 0.460 & 0.455 & 0.946 & 0.774 & 0.467 & 0.448 & \underline{0.444} & \underline{0.445} & 0.457 & 0.449 & 0.460 & 0.457 \\
ETTh2 & \textbf{0.371} & \textbf{0.400} & 0.450 & 0.461 & 0.382 & 0.410 & 0.407 & 0.421 & 4.522 & 1.696 & 0.393 & 0.411 & \underline{0.378} & \underline{0.401} & 0.383 & 0.407 & 0.564 & 0.519 \\
ETTm1 & \textbf{0.374} & \textbf{0.395} & 0.578 & 0.515 & 0.388 & 0.402 & 0.411 & 0.418 & 0.929 & 0.730 & \underline{0.381} & \underline{0.397} & 0.382 & \underline{0.397} & 0.407 & 0.411 & 0.404 & 0.408 \\
ETTm2 & 0.280 & 0.328 & 0.333 & 0.370 & 0.291 & 0.335 & 0.295 & 0.332 & 1.553 & 0.889 & \underline{0.276} & \underline{0.323} & \textbf{0.274} & \textbf{0.321} & 0.291 & 0.334 & 0.355 & 0.402 \\
Weather & 0.245 & \underline{0.274} & 0.358 & 0.392 & 0.259 & 0.281 & 0.259 & 0.286 & 0.625 & 0.557 & \underline{0.244} & \underline{0.274} & \textbf{0.241} & \textbf{0.271} & 0.260 & 0.281 & 0.265 & 0.317 \\
Exchange & 0.369 & \textbf{0.408} & 0.505 & 0.499 & 0.396 & 0.421 & 0.406 & 0.438 & 1.297 & 0.866 & 0.388 & 0.418 & 0.387 & 0.417 & \underline{0.364} & \textbf{0.408} & \textbf{0.340} & \underline{0.414} \\
ILI & \textbf{1.999} & 0.929 & 3.144 & 1.210 & \underline{2.165} & \textbf{0.899} & 2.241 & \underline{0.923} & 4.744 & 1.462 & 2.581 & 1.031 & 2.577 & 1.008 & 2.964 & 1.168 & 4.455 & 1.554 \\
ECL & 0.186 & 0.275 & 0.285 & 0.367 & 0.204 & 0.294 & 0.197 & 0.297 & 0.409 & 0.473 & 0.185 & \underline{0.274} & \textbf{0.171} & \textbf{0.270} & \underline{0.180} & \textbf{0.270} & 0.225 & 0.319 \\
Traffic & 0.501 & 0.315 & 0.667 & 0.416 & 0.482 & 0.308 & 0.626 & 0.332 & 0.667 & 0.366 & 0.512 & 0.307 & \underline{0.466} & \underline{0.287} & \textbf{0.423} & \textbf{0.282} & 0.673 & 0.419 \\
\midrule
\textbf{Avg.\ rank (MSE)} & \multicolumn{2}{c}{2.33} & \multicolumn{2}{c}{7.67} & \multicolumn{2}{c}{4.00} & \multicolumn{2}{c}{5.44} & \multicolumn{2}{c}{8.78} & \multicolumn{2}{c}{4.00} & \multicolumn{2}{c}{2.22} & \multicolumn{2}{c}{4.00} & \multicolumn{2}{c}{6.56} \\
\bottomrule
\end{tabular}}
\end{table}

Among the models of Table 1, ours has the lowest MSE on ETTh1, ETTh2, ETTm1 and ILI, and improves on PatchTST on eight of the nine datasets, by $4.6\%$ on average over all nine and on TimesNet by $9.1\%$. TimeXer is the strongest baseline and the two are level, with average ranks of 2.33 and 2.22, and they are the only models never ranked worse than fourth on any dataset. Ours is behind TimeXer on ETTm2 and Weather by under $0.007$ MSE, and by more on ECL and Traffic, discussed next.

On ECL and Traffic we trail TimeXer and iTransformer, both of which use information from the other channels when predicting each one, whereas our design is channel-independent. TimeXer's own ablation reports that on Traffic, attending over all channels together outperforms its proposed design~\citep{wang2024timexer}. On ECL we still beat PatchTST by $8.8\%$ (0.186 vs.\ 0.204). On Traffic we do not, 0.501 against its 0.482. PatchTST is channel-independent like ours, so cross-channel information cannot explain that gap. §5.1 shows that on Traffic, learning the filter makes the forecast $2.7\%$ worse.

Over the nine datasets our model is significantly better than six of the eight baselines (paired Wilcoxon signed-rank on per-dataset MSE, $p<0.1$, Autoformer, TimesNet and Transformer 9-0, DLinear 8-1, PatchTST 8-1, TimeMixer 6-3). iTransformer (6-3, $p=0.20$) and TimeXer (5-4, $p=0.50$) are statistically indistinguishable from ours, see Appendix~\ref{app:significance}. Full per-horizon results are in Appendix~\ref{app:perhorizon}. §5 analyses how the margin depends on tuning and how structure and learning decouple.

Table~\ref{tab:dynpatch} compares the four dynamic-patching methods, with per-horizon results in Table~\ref{tab:dynpatch_h}. Ours has the lowest MSE on ETTh1, ETTm1 and Weather and is ahead of EntroPE on all seven datasets. Pathformer is the strongest of the four and leads on ETTh2, ETTm2 and ECL. Every model is the mean over the same three seeds, whereas EntroPE's paper reports the best of five. The within-model control of §5.3 compares patchers with every other factor held fixed. It is the test \citet{zucchi2026} apply to EntroPE, TimeMosaic and HDMixer, where a tuned uniform grid beats them on 38\% to 72\% of settings against 25\% for ours.

\begin{table}[h!]\centering\small\setlength{\tabcolsep}{3pt}
\caption{Dynamic-patching methods: MSE / MAE averaged over horizons, lookback $L{=}96$. Best in \textbf{bold}, second best \underline{underlined}. Every model is run in our harness from its released code and published configuration and is the per-task mean over seeds \{2021, 42, 123\}. Per-dataset standard deviations are in Table~\ref{tab:dynpatch_sd}. TimeMosaic and HDMixer run their released scripts with the lookback set to 96. HDMixer releases configurations only for ETTh1 and ETTh2. Pathformer diverged on Weather and its Traffic runs did not complete.}
\label{tab:dynpatch}
\resizebox{\textwidth}{!}{%
\begin{tabular}{lcccccccccc}
\toprule
Dataset & \multicolumn{2}{c}{Ours} & \multicolumn{2}{c}{EntroPE} & \multicolumn{2}{c}{Pathformer} & \multicolumn{2}{c}{TimeMosaic} & \multicolumn{2}{c}{HDMixer} \\
 & MSE & MAE & MSE & MAE & MSE & MAE & MSE & MAE & MSE & MAE \\ \midrule
ETTh1 & \textbf{0.421} & 0.438 & \underline{0.438} & 0.437 & 0.445 & \textbf{0.426} & 0.450 & 0.435 & 0.441 & \underline{0.433} \\
ETTh2 & 0.371 & 0.400 & 0.381 & 0.409 & \textbf{0.346} & \textbf{0.379} & 0.373 & 0.394 & \underline{0.353} & \underline{0.387} \\
ETTm1 & \textbf{0.374} & 0.395 & 0.386 & 0.397 & \underline{0.381} & \underline{0.382} & 0.382 & \textbf{0.380} & -- & -- \\
ETTm2 & 0.280 & 0.328 & 0.294 & 0.338 & \textbf{0.273} & \textbf{0.316} & \underline{0.278} & \underline{0.317} & -- & -- \\
Weather & \textbf{0.245} & \underline{0.274} & \underline{0.249} & 0.279 & -- & -- & 0.252 & \textbf{0.267} & -- & -- \\
ECL & \underline{0.186} & 0.275 & 0.196 & 0.288 & \textbf{0.170} & \textbf{0.260} & 0.190 & \underline{0.274} & -- & -- \\
Traffic & \underline{0.501} & 0.315 & 0.513 & \underline{0.303} & -- & -- & \textbf{0.454} & \textbf{0.278} & -- & -- \\
\bottomrule
\end{tabular}}
\end{table}

\newpage
\section{Analysis and Discussion}

The main contribution of this paper is to separate three coupled choices that prior learnable-wavelet work usually credits with a single improvement number, the use of a multi-resolution decomposition at all, the choice of starting wavelet and the training of the filter. We do so with controlled ablations at the final tuned configuration and ask three questions in turn. The competitive benchmark of §\ref{results} provides supporting evidence. Tables~\ref{tab:main} and \ref{tab:mode_ladder} report three-seed means. Every control in this section holds both arms at seed 2021, so its numbers for the main model differ slightly from those means.

\subsection{Isolating decomposition, filter learning and dynamic boundaries}

We isolate each factor with a mode ladder (Table~\ref{tab:mode_ladder}, Table~\ref{tab:ladder_seed}, in which each mode adds exactly one capability over the previous, with everything else held at the task's tuned configuration:

\begin{itemize}
    \item none/uniform vs PatchTST. Is the gain the backbone rather than the wavelet?
    \item Fixed/uniform vs none/uniform. Does a frozen decomposition help, with the patching held uniform?
    \item Fixed/merged vs Fixed/uniform. Does merging windows into variable-length patches help?
    \item Learned/merged vs Learned/uniform. Does merging help once the filter is trained?
    \item Learned vs Fixed. Does adapting the filter help beyond a good fixed wavelet?
    \item Dynamic vs Learned. Does recomputing the patch boundaries each step help?
\end{itemize}

\begin{table}[h!]
  \centering
  \begin{threeparttable}
  \caption{Mode ladder: test MSE averaged over the four horizons. PatchTST (no decomposition), then the three DWT modes of our model: Fixed (frozen filter), Learned (trainable filter, cached boundaries, the main model) and Dynamic (boundaries recomputed each step, straight-through band embedding of \S3.5). Best per row in \textbf{bold}.}
  \label{tab:mode_ladder}
  \begin{tabular*}{\textwidth}{@{\extracolsep{\fill}}lcccc@{}}
    \toprule    
    Dataset & PatchTST & Fixed & Learned & Dynamic \\
    \midrule
    ETTh1    & 0.449 & \textbf{0.421} & \textbf{0.421} & \textbf{0.421} \\
    ETTh2    & 0.382 & \textbf{0.371} & \textbf{0.371} & 0.373 \\
    ETTm1    & 0.388 & 0.379 & 0.374 & \textbf{0.373} \\
    ETTm2    & 0.291 & 0.283 & 0.280 & \textbf{0.279} \\
    Weather  & 0.259 & 0.255 & \textbf{0.245} & \textbf{0.245} \\
    Exchange & 0.396 & \textbf{0.366} & 0.369 & 0.379 \\
    ILI      & 2.165 & \textbf{1.996} & 1.999 & 2.011 \\
    ECL      & 0.204 & 0.197 & \textbf{0.186} & \textbf{0.186} \\
    Traffic  & \textbf{0.482} & 0.488 & 0.501 & 0.501 \\
    \bottomrule
  \end{tabular*}
  \end{threeparttable}
\end{table}

\begin{table}[t]
\centering\small
\setlength{\tabcolsep}{4pt}
\caption{Within-architecture ladder. Four-horizon mean test MSE, seed 2021. none/uniform keeps the whole backbone, the pooled patch tokens, the cross-attention and the head, with the raw series as its single scale token, no decomposition and one patch per base window, Fixed/uniform adds the frozen DWT decomposition with the patching still uniform, Fixed/merged merges same-band windows. Learned/uni-16 and Learned/uni-$b$ keep the trained filter and replace merging with a fixed grid, of 16 steps and of the task's tuned base width $b$. Learned/merged trains the filter with cached boundaries (the main model) and Dynamic recomputes the boundaries every step. Best per row in \textbf{bold}.}
\label{tab:ladder_seed}
\begin{tabular}{lcccccccc}
\toprule
Dataset & PatchTST & none & Fixed & Fixed & Learned & Learned & Learned & Dynamic \\
        &          & uniform & uniform & merged & uni-16 & uni-$b$ & merged & \\
\midrule
ETTh1    & 0.449 & 0.421 & \textbf{0.418} & \textbf{0.418} & 0.424 & 0.419 & \textbf{0.418} & \textbf{0.418} \\
ETTh2    & 0.382 & 0.375 & 0.371 & 0.369 & 0.380 & \textbf{0.368} & \textbf{0.368} & 0.370 \\
ETTm1    & 0.388 & 0.383 & 0.379 & 0.379 & 0.375 & 0.374 & 0.373 & \textbf{0.372} \\
ETTm2    & 0.291 & 0.283 & 0.280 & 0.281 & 0.289 & \textbf{0.277} & \textbf{0.277} & \textbf{0.277} \\
Weather  & 0.259 & 0.258 & 0.255 & 0.255 & 0.246 & 0.246 & \textbf{0.245} & \textbf{0.245} \\
Exchange & 0.396 & 0.378 & 0.360 & 0.360 & 0.372 & 0.357 & \textbf{0.354} & 0.376 \\
ILI      & 2.165 & 2.246 & 2.009 & 1.968 & 2.094 & 2.016 & 1.975 & \textbf{1.958} \\
ECL      & 0.204 & 0.199 & 0.196 & 0.197 & \textbf{0.186} & \textbf{0.186} & \textbf{0.186} & \textbf{0.186} \\
Traffic  & \textbf{0.482} & 0.490 & 0.486 & 0.488 & 0.511 & 0.498 & 0.500 & 0.500 \\
\midrule
Mean vs PatchTST & & $-1.6\%$ & $-4.0\%$ & $-4.2\%$ & $-3.1\%$ & $-5.1\%$ & $-5.4\%$ & $-4.8\%$ \\
\bottomrule
\end{tabular}
\end{table}

\textbf{Does the decomposition help?} A frozen wavelet, with no learning at all, improves on PatchTST on eight of the nine datasets, by 1.5\% to 7.8\% (largest on ILI 7.8\%, Exchange 7.6\% and ETTh1 6.2\%). Traffic is the one exception, where the frozen decomposition is 1.2\% worse. This gain comes from the multi-resolution structure alone, since the Fixed mode involves no filter training. Part of the gap to PatchTST is the backbone. With no decomposition at all, our backbone with uniform patches already improves on PatchTST on seven of the nine datasets, and adding the frozen decomposition to it, with the patching still uniform, improves all nine, by 0.6\% to 1.4\% on six datasets, 4.8\% on Exchange, 10.5\% on ILI and 0.9\% on Traffic (Table~\ref{tab:ladder_seed}). Merging the windows with the filter frozen then changes the MSE by less than 0.5\% on seven datasets, improves ETTh2 by 0.7\% and ILI by 2.1\%, and costs Traffic 0.4\%.

\textbf{Does merging help once the filter is trained?} Yes. In Learned mode, merging beats a fixed grid of 16 steps on eight of the nine datasets, by 2.2\% on average, and is level with a grid at the tuned base width while using 48\% to 63\% fewer patches.

\textbf{Does learning the filter help?} Only on some datasets and the pattern is not the one we expected. The two largest effects are ECL (5.6\%) and Weather (3.9\%), followed by ETTm1 (1.3\%). On ETTm2 (1.1\%), ETTh1, ETTh2, Exchange and ILI the change is within the seed-to-seed spread of the model, so we read it as no measurable effect. On Traffic learning the filter makes the forecast 2.7\% worse. Filter learning therefore contributes between nothing and most of the total gain depending on the dataset. The nine-dataset mean is a 0.9\% improvement, spanning a 5.6\% gain on one dataset and a 2.7\% loss on another. §5.4 asks which datasets it helps and why.

\textbf{Does recomputing the boundaries help?} No. Over the 36 tasks Dynamic is better than Learned on 18 and worse on 18, and 16 of the differences lie inside 0.0005. By dataset, Dynamic matches Learned to within 0.4\% on eight datasets and is 2.8\% worse on Exchange. Recomputing the patch boundaries from the current filter at every step is no better than computing them once and caching them.

\subsection{Initialisation and the role of filter adaptation}

Two experiments test whether the filter itself matters. Initialisation (wavelet vs. Xavier vs. Random) asks whether a good wavelet prior is needed. The Learned vs Fixed comparison of §5.1 asks whether training the filter adds anything on top.

\newcolumntype{C}{>{\centering\arraybackslash}X}
\begin{table}[h!]
\centering
\begin{threeparttable}
\caption{Initialisation sensitivity: test MSE for the learnable filter (Learned mode, the main model) started from three initialisations. Wavelet is each dataset's tuned wavelet (db8; haar for Exchange); Xavier and Random are non-wavelet filters of the same width. Single horizon (p96; ILI p24). Best per row in \textbf{bold}.}
\label{tab:init_sensitivity}
\begin{tabularx}{\textwidth}{l C C C}
\toprule
Dataset & Wavelet & Xavier & Random \\
\midrule
ETTh1    & \textbf{0.365} & 0.382 & 0.383 \\
ETTh2    & \textbf{0.281} & 0.305 & 0.297 \\
ETTm1    & \textbf{0.313} & 0.323 & 0.318 \\
ETTm2    & \textbf{0.173} & 0.178 & 0.181 \\
Weather  & \textbf{0.163} & 0.164 & \textbf{0.163} \\
Exchange & \textbf{0.082} & 0.088 & 0.086 \\
ILI      & \textbf{1.876} & 2.931 & 2.666 \\
ECL      & \textbf{0.159} & 0.174 & 0.178 \\
Traffic  & \textbf{0.474} & 0.493 & NaN\tnote{$\dagger$} \\
\bottomrule
\end{tabularx}
\begin{tablenotes}[flushleft]
\footnotesize
\item $^{\dagger}$ Random diverged to NaN on Traffic.
\end{tablenotes}
\end{threeparttable}
\end{table}

Table~\ref{tab:init_sensitivity} reports the effects. A wavelet start is best or level on all nine datasets and its advantage over a non-wavelet start ranges from nothing to very large. On Weather the three starts are within 0.4\% of each other. On ILI, a non-wavelet start costs 42 to 56\%. With only 966 timesteps and a lookback of 36, the model likely has too little data to recover from a poor decomposition, so the wavelet start supplies structure that larger datasets can learn on their own.

A valid multi-resolution start is necessary on every dataset, while the specific wavelet within that family is not (§5.5), so the two are separate properties. On Weather any filter works, the three starts are within 0.4\%, whereas on ECL a non-wavelet start costs 9.3\% to 11.7\%, yet both are datasets where learning the filter helps most.

Why the learnable filter changes the forecast so little is a question about the architectural rather than the tuning, and §5.6 addresses it with the measured gradient flow.

\subsection{A within-model control for dynamic vs uniform patching}

We test our model against itself with the merge step switched off (Table~\ref{tab:uniform}). Ours-Learned merges same-scale base windows into variable-length patches. Ours-Uniform keeps every base window as one fixed-size patch, with the patch size $b$ selected on validation per dataset and horizon from $\{4, 8, 16, 24, 32\}$. Everything else (the learnable DWT, backbone, head, optimiser and training budget) is held identical, so the patcher is the only difference. Both arms share one configuration and one harness, so the comparison isolates adaptive patching more cleanly than any comparison across published models.

\begin{table}[h!]\centering\small
\caption{Within-model uniform control (Learned mode, seed 2021). Only the patcher differs. MSE averaged over four horizons; $\Delta$MSE $<0$ favours dynamic.}
\label{tab:uniform}
\begin{tabular*}{\textwidth}{@{\extracolsep{\fill}}lcccc@{}}\toprule
Dataset & Ours-Learned MSE & Ours-Uniform MSE & $\Delta$MSE & Horizons won \\ \midrule
ETTh1    & 0.418 & 0.432 & $-3.2\%$  & 4/4 \\
ETTh2    & 0.368 & 0.382 & $-3.5\%$  & 4/4 \\
ETTm1    & 0.373 & 0.373 & $-0.0\%$  & 1/4 \\
ETTm2    & 0.277 & 0.280 & $-1.0\%$  & 4/4 \\
Weather  & 0.245 & 0.246 & $-0.4\%$  & 3/4 \\
Exchange & 0.354 & 0.397 & $-10.8\%$ & 3/4 \\
ILI      & 1.975 & 2.180 & $-9.4\%$  & 4/4 \\
ECL      & 0.186 & 0.186 & $+0.2\%$  & 1/4 \\
Traffic  & 0.500 & 0.502 & $-0.3\%$  & 3/4 \\
\bottomrule\end{tabular*}\end{table}

Variable-length merging wins or ties on eight of the nine datasets, by up to $10.8\%$ on Exchange. ECL is the one dataset where the tuned uniform grid wins, by $0.2\%$ on average. Overall a tuned uniform grid matches or beats variable-length merging on 9 of 36 settings ($25.0\%$).

The mode ladder of §5.1 and this control together show that the two components help on different datasets. On ECL, Weather and ETTm1 the gain comes from the filter and variable-length merging has no effect. On ETTh1 and ETTh2 the gain comes from the patcher and the filter has no effect. ETTm2 and Exchange gain from merging alone. The two are not substitutes, so a single averaged improvement number combines two separate effects into one figure. §5.4 examines which datasets show the filter gain.

This within-model control is the test \citet{zucchi2026} use to argue that adaptive patching rarely improves on a tuned uniform grid. Over their 8 datasets and 4 horizons a tuned uniform grid beats the three published adaptive patchers on 56.3\% (EntroPE), 71.9\% (HDMixer) and 37.5\% (TimeMosaic) of settings. Following their protocol, with the uniform patch size selected on validation from the same grid, the grid beats ours on 25.0\% of our 36 settings, and on 28.1\% of their exact 32 (their panel is ours without ILI). Our variable-length patches therefore hold up better against a tuned uniform grid than any of the three adaptive patchers they test. The three methods in their comparison place boundaries from time-domain signals, predictive entropy in EntroPE, temporal-heterogeneity routing in TimeMosaic and a learned deformable patcher in HDMixer, while ours places them from per-scale wavelet energy.
 
\textbf{Do the cut positions matter?} Under Dynamic the cuts move during training and the forecast does not change much (\S5.6). We push this further with a random-merging control, a test we did not find in the dynamic patching methods of Table~\ref{tab:dynpatch}. It keeps the Learned model and the number of patches the DWT assigns to every window and channel, and draws the cut positions uniformly at random among the window edges. This moves 53\% to 73\% of the cuts while leaving the patch count, the mean patch length and the length distribution unchanged (Appendix~\ref{app:patchcounts}). Table~\ref{tab:cuts} compares it with our patcher and with a non-overlapping uniform grid of 16 steps, PatchTST's patch length, all at seed 2021. Random cuts forecast almost as well as ours, within 1\% on seven of the nine datasets. The 16-step uniform grid is worse than ours on every dataset, by 0.4\% to 6.0\%. The random control still takes the patch count of every window and the band tags from the DWT. What the patcher delivers is therefore the decomposition, its band tags and the count per window, with 48\% to 63\% fewer tokens, and in this architecture the exact cut positions add little.

\begin{table}[h!]
\centering\small
\caption{Cut positions and token count (Learned mode, seed 2021). Four-horizon mean test MSE for our patcher, a non-overlapping uniform grid of 16 steps and random merging, which keeps our patch count per window and channel and so our mean patch length. Patches are per window and channel on the training split, length in time steps. Best per row in \textbf{bold}.}
\label{tab:cuts}
\begin{tabular*}{\textwidth}{@{\extracolsep{\fill}}lccccccc@{}}\toprule
 & \multicolumn{3}{c}{MSE} & \multicolumn{2}{c}{Patches} & \multicolumn{2}{c}{Mean length} \\
\cmidrule(lr){2-4}\cmidrule(lr){5-6}\cmidrule(lr){7-8}
Dataset & Ours & Uniform 16 & Random & Ours & Uniform 16 & Ours & Uniform 16 \\ \midrule
ETTh1    & \textbf{0.418} & 0.424 & \textbf{0.418} & 4.3  & 6 & 24.2 & 16 \\
ETTh2    & \textbf{0.368} & 0.380 & 0.372 & 14.0 & 6 & 8.4  & 16 \\
ETTm1    & \textbf{0.373} & 0.375 & 0.374 & 13.8 & 6 & 13.3 & 16 \\
ETTm2    & \textbf{0.277} & 0.289 & 0.279 & 19.8 & 6 & 4.9  & 16 \\
Weather  & \textbf{0.245} & 0.246 & 0.246 & 6.0  & 6 & 16.7 & 16 \\
Exchange & \textbf{0.354} & 0.372 & 0.376 & 13.2 & 6 & 13.5 & 16 \\
ILI      & \textbf{1.975} & 2.094 & 2.255 & 5.9  & 2 & 6.2  & 16 \\
ECL      & \textbf{0.186} & \textbf{0.186} & \textbf{0.186} & 10.6 & 6 & 23.2 & 16 \\
Traffic  & \textbf{0.500} & 0.511 & 0.502 & 16.5 & 6 & 6.3  & 16 \\
\bottomrule\end{tabular*}\end{table}

\subsection{Dataset properties governing the improvement}

The structural gain of §5.1 appears on eight of the nine datasets, the learnable-filter gain on only three. This mixes two questions, how far the filter moves and whether that movement improves the forecast. Treated as one they reduce to ``bigger datasets help", so we separate them and then ask which datasets improve.

\textbf{How far the filter moves is set by data volume.} We truncate a single series (ETTm1, with ETTm2 mirroring it) so that \texttt{$d_{model}$}, filter width, wavelet, channel count, domain and spectrum are all frozen and only the number of training rows changes (Table~\ref{tab:truncation}, Test~A). Filter drift falls monotonically as rows shrink, from 0.082 at the full arm to 0.020 at the $\tfrac{1}{16}$ arm. At the $\tfrac{1}{4}$ arm (8{,}640 rows, the size of the ETTh sets) drift is 0.050, almost exactly the ETTh sets its own drift at that row count (ETTh1 0.049, ETTh2 0.044). Much of the cross-dataset drift range is reproduced inside one series by changing nothing but the row count. Holding the row count fixed at 8,640 and switching 15-min to hourly (Test B), which also changes the calendar span the data covers, leaves drift essentially unchanged (a move of only 0.003 to 0.008, with opposite signs), an order of magnitude smaller than the volume swing. Drift magnitude is thus set by the number of training rows, not by the sampling resolution or the calendar span they cover. 

The truncation test holds channel count fixed, so it cannot rule out channel count as a driver of drift. ECL and Traffic can. They have 321 and 862 channels against ETT's 7, and their training-row counts fall between the ETTh and ETTm sets. Their drift falls there too, at 0.065 and 0.067. Sorting the nine datasets by training rows gives almost the same order as sorting them by drift, with Weather the only one out of place. Channel count therefore does not set drift magnitude either.

\begin{table}[h!]
\centering
\begin{threeparttable}
\caption{Truncation control (Learned Mode, the main model, 10 epochs / patience 3 at $H=96$). Only the training-row count changes, with \texttt{$d_{model}$}, wavelet, filter width, channels and spectrum frozen.  \emph{Test A} shrinks ETTm1's rows at fixed 15-min sampling (isolates volume). \emph{Test B} holds 8{,}640 rows and switches 15-min $\to$ hourly (isolates resolution). Drift is filter movement from init for the low-pass $h$ and high-pass $g$}
\label{tab:truncation}
\begin{tabular*}{\textwidth}{@{\extracolsep{\fill}}lrccc@{}}
\toprule
\multicolumn{5}{@{}l}{\emph{Test A --- shrink rows, sampling fixed (isolates volume)}}\\
\midrule
Arm & Train rows & Drift $h$ & Drift $g$ & Test MSE \\
\midrule
full                       & 34{,}560 & 0.082 & 0.116 & 0.313 \\
$\tfrac{1}{2}$             & 17{,}280 & 0.057 & 0.057 & 0.392 \\
$\tfrac{1}{4}$ (ETTh size) & 8{,}640  & 0.050 & 0.036 & 0.447 \\
$\tfrac{1}{8}$             & 4{,}320  & 0.038 & 0.015 & 0.578 \\
$\tfrac{1}{16}$            & 2{,}160  & 0.020 & 0.013 & 0.826 \\
\bottomrule
\end{tabular*}

\vspace{0.7em}

\begin{tabular*}{\textwidth}{@{\extracolsep{\fill}}lccc@{}}
\toprule
\multicolumn{4}{@{}l}{\emph{Test B --- hold 8{,}640 rows, 15-min $\to$ hourly (isolates resolution)}}\\
\midrule
Series & 15-min $h$-drift & hourly $h$-drift & $\Delta$ \\
\midrule
ETTm1 & 0.050 & 0.047 & $-0.003$ \\
ETTm2 & 0.036 & 0.044 & $+0.008$ \\
\bottomrule
\end{tabular*}

\begin{tablenotes}[flushleft]
\footnotesize
\item ETTm2 mirrors ETTm1 (full $0.059 \to \tfrac{1}{4}$ data $0.036$).
Cross-dataset drift values for comparison (§5.6) are ILI 0.007, Exchange 0.019, ETTh2 0.044, ETTh1 0.049, Traffic 0.065, ECL 0.067, Weather 0.069, ETTm2 0.070 and ETTm1 0.078.\end{tablenotes}
\end{threeparttable}
\end{table}

The gain appears on two of the three series that contain fast fluctuations, Weather (10-min) and ETTm1 (15-min), and ETTm2's $-1.1\%$ sits just inside its noise band. On all three finely sampled datasets the DWT's high-pass filter, which extracts the fast detail band, reshapes more during training than its low-pass counterpart, with $g/h$ of $1.33$ on Weather, $1.43$ on ETTm2 and $1.65$ on ETTm1 (§5.6). The coarsely sampled Exchange (daily) and ILI (weekly) contain no fast fluctuations for the high-pass to adapt to, so learning has no effect there no matter how large the dataset.

Fine sampling is therefore enough to produce the gain, but it is not required. Hourly ECL gains the most of any dataset at $5.6\%$ and we cannot explain why. The high-pass explanation does not cover ECL, because there the two filters move by almost the same amount 0.93 and ETTh1 shows the same ratio at 0.92 while gaining nothing. Traffic is the opposite case. It is hourly like ECL, its filter moves just as far and its ratio of $1.24$ sits close to Weather's, yet learning the filter makes the forecast $2.7\%$ worse. Across all nine datasets the ratio does not order the gain, so we read it as a description of how the finely sampled series adapt rather than a predictor of which datasets improve. Of the four hourly datasets one gains, two are unaffected and one is harmed and we cannot separate them. In summary, volume sets how far the filter moves and fast content is enough for that movement to help, but the hourly datasets remain unexplained.

A domain-controlled version of this test is already built into the benchmark. ETTh1/ETTm1 and ETTh2/ETTm2 are the same two electricity substations recorded at hourly and 15-minute resolution, so within each pair the signal, domain, channel count and calendar span are identical. Only the sampling rate changes and with it the training-row count (15-min gives roughly $4\times$ the rows). The truncation control above isolates that volume difference, showing it sets drift magnitude, not whether drift helps. The learnable filter helps on ETTm1 ($-1.3\%$), sits at the edge of noise on ETTm2 ($-1.1\%$) and helps on neither hourly series (ETTh1 $0.0\%$, ETTh2 $0.0\%$). Within this matched pair the finer sampling enables the gain by exposing fluctuations that hourly sampling cannot resolve. ECL shows that a gain can occur without fine sampling, though we cannot say what causes it there.

\subsection{Robustness to structural choices}

The model is robust to its structural hyperparameters. Varying wavelet family (ten wavelets, 2 to 40 taps), decomposition level $\{2,4,6,8\}$ and base-patch size $\{4,8,16,24\}$ one at a time, with everything else at the task-tuned configuration, moves test MSE by only a fraction of a percent on most datasets mostly within the seed-noise band of §5.1. So none of these is a strong optimum (full results in Appendix~\ref{app:sweeps}). The Daubechies width curve is nearly flat with its minimum at db8 (16 taps), our default. Decomposition level is the only mildly data-specific axis. The pure-trend Exchange prefers the shallowest level, ETTh2 and ILI a deeper one, and the rest are indifferent (level CV under $1\%$).

\subsection{Mechanistic interpretation of the main framework}

A dedicated mechanistic run instruments the main model on all nine datasets. Each epoch we track the same set of training windows per dataset and record how far the learned filter of each channel has moved from its wavelet initialisation (Filter drift), how the energy of the decomposition is spread across scales (Scale energy shift) and the size of the gradient reaching the filter relative to the gradient on the rest of the model at the final epoch (Gradient share). Table~\ref{tab:mechanism} reports the Learned mode, and the same statistics for Dynamic are in Appendix~\ref{app:mechanism}.

\begin{table}[h!]
\centering
\begin{threeparttable}
\caption{Final-epoch mechanism statistics (Learned), averaged over the four horizons, ordered by scale energy shift. Scale energy shift is measured on a set of training windows tracked through training. High-pass drift is omitted since it equals drift $h$ times the ratio $g/h$. The full set for both modes is in Appendix~\ref{app:mechanism}. The last column gives the change in MSE from freezing the filter to learning it (§5.1), with negative meaning learning helps.}
\label{tab:mechanism}
\begin{tabular}{lccccc}
\toprule
Dataset & Drift $h$ & $g/h$ & Grad share & Scale energy shift (\%) & Learnable helps? \\
\midrule
ETTm2    & 0.070 & 1.43 & 0.78 & 15.9 & yes, $-1.1\%$ \\
Weather  & 0.069 & 1.33 & 1.08 & 11.5 & yes, $-3.9\%$ \\
ETTh2    & 0.044 & 1.20 & 0.87 & 9.0  & no, $0.0\%$ \\
ETTm1    & 0.078 & 1.65 & 0.43 & 8.9  & yes, $-1.3\%$ \\
ETTh1    & 0.049 & 0.92 & 0.26 & 2.7  & no, $0.0\%$ \\
ILI      & 0.007 & 0.89 & 0.46 & 2.5  & no, $+0.2\%$ \\
Traffic  & 0.065 & 1.24 & 0.09 & 2.1  & no, $+2.7\%$ \\
ECL      & 0.067 & 0.93 & 0.14 & 1.6  & yes, $-5.6\%$ \\
Exchange & 0.019 & 1.26 & 0.57 & 0.8  & no, $+0.8\%$ \\
\bottomrule
\end{tabular}
\end{threeparttable}
\end{table}

\begin{figure}[h!]
  \centering
  \includegraphics[width=\linewidth]{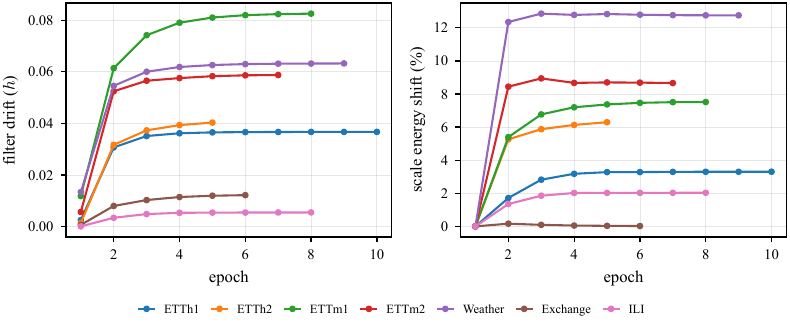}
  \caption{Adaptation settles by epoch three. Filter drift (left) and scale energy shift (right), Learned mode, 96-step horizon (ILI 24). Table \ref{tab:mechanism} reports the same statistics averaged over the four horizons, so its values differ from the right edge here. The grouping of datasets is the same in both.}
  \label{fig:mech_dynamics}
\end{figure}

Adaptation happens early and then stops. Filter drift and scale energy climb over the first two or three epochs and then plateau (Figure \ref{fig:mech_dynamics}). Where the filter drifts, the dynamic mode recomputes cuts that differ from the fixed wavelet cuts the cached mode keeps, yet the two still match in MSE (§5.1). The cuts do move. On the tracked windows, between $8\%$ (Exchange, ILI) and $84\%$ (ETTm1) of boundaries sit at a different position at the final epoch than at epoch one. The partition is neutral because moving them does not change the forecast.

A direct ablation tests whether the boundary path contributes to the forecast. The gradient to the filter enters the patching only through the band embedding of §3.5, a straight-through estimator (STE). Its forward value is the hard embedding of the patch's dominant band. Its gradient flows through the soft embedding, a soft-weight mixture of the band rows at strength $\alpha$, which the deployed configuration sets to one. The path is active only in the Dynamic mode, so the ablation runs there. At $\alpha=0$ the estimator reduces to the pure argmax, so no gradient reaches the filter through the patching and the forecast barely changes. Averaged over the nine datasets the MSE is $0.9\%$ lower with the gradient on (Table~\ref{tab:hybrid_off}), and $0.3\%$ lower without ILI. Only two datasets move by more than 1\%, ETTh2 ($-1.6\%$) and ILI ($-5.2\%$), both in favour of the gradient. So the gradient that lets the patching train the filter has at most a small effect, and together with the Learned versus Dynamic tie of §5.1 this shows the boundaries are close to neutral whether or not they move and whether or not their gradient flows. The filter keeps learning with this path cut because it has two roles in the model and only one was cut. Besides tagging each patch with a band, the filter decomposes the signal into the multi-resolution reconstruction the transformer reads as scale tokens. That path has no argmax, so it passes a gradient to the filter regardless of the tagging.

\begin{table}[h!]
\centering
\caption{Boundary-gradient ablation (Dynamic mode). Four-horizon mean test MSE, straight-through strength $\alpha$ on (main) vs off. $\Delta$ is $\alpha{=}1$ relative to $\alpha{=}0$, so $\Delta<0$ favours the gradient.}
\label{tab:hybrid_off}
\begin{tabular}{lccc}
\toprule
Dataset & $\alpha{=}1$ (main) & $\alpha{=}0$ & $\Delta$ (\%) \\
\midrule
ETTh1    & 0.4178 & 0.4211 & $-0.78$ \\
ETTh2    & 0.3704 & 0.3763 & $-1.56$ \\
ETTm1    & 0.3724 & 0.3726 & $-0.05$ \\
ETTm2    & 0.2770 & 0.2773 & $-0.13$ \\
Weather  & 0.2453 & 0.2452 & $+0.02$ \\
Exchange & 0.3761 & 0.3765 & $-0.10$ \\
ILI      & 1.9582 & 2.0664 & $-5.24$ \\
ECL      & 0.1862 & 0.1862 & $+0.00$ \\
Traffic  & 0.5003 & 0.5001 & $+0.05$ \\
\midrule
mean     &        &        & $-0.86$ \\
\bottomrule
\end{tabular}
\end{table}

Learning reshapes the scale decomposition on the finely sampled datasets, where the energy redistribution reaches 9\% to 16\%, but the size of that reshaping does not track the gain. ECL redistributes almost the least, 1.6\% against Exchange's 0.8\%, and gains the most, while ETTh2 redistributes 9\% for no measurable gain. The filter can therefore change the decomposition substantially without improving MSE, and can improve MSE while barely changing it. The scale energy in Table \ref{tab:mechanism} measures how much the model redistributes energy across scales during training. We average the per-band energy over the tracked sequences at the first and last epoch and report the total variation distance between the two. These windows overlap and cover only the start of the training series, so the statistic is coarse and we read it qualitatively. It is large on the three fast content datasets (ETTm2 $15.9\%$, Weather $11.5\%$, ETTm1 $8.9\%$), two of which benefit, equally large on ETTh2 ($9.0\%$) which does not benefit.

On the three finely sampled datasets, two of which benefit, training moves the high-pass filter $g$ further than the low-pass $h$, with $g/h$ of $1.33$ on Weather, $1.43$ on ETTm2 and $1.65$ on ETTm1, consistent with the filter adapting its high-frequency response. The sampling rate caps the fastest content a series can record (the Nyquist limit, half the sampling rate), so daily Exchange and weekly ILI contain no fast content for the high-pass to adapt to, and neither gains (Table~\ref{tab:freq}). Excluding ECL and Traffic, the three finely sampled series have the highest $g/h$ in both modes (Table~\ref{tab:mechanism_full}). The ratio does not, however, separate the datasets where learning helps from those where it does not. ETTm2 sits at $1.43$ without a measurable gain. ECL and Traffic break the pattern in opposite directions. ECL gains the most of any dataset at a ratio of $0.93$, below every other dataset that benefits, while on Traffic, at $1.24$ and $1.35$ in Dynamic (Table~\ref{tab:mechanism_full}), learning the filter makes the forecast $2.7\%$ worse. We therefore read $g/h$ as describing how the finely sampled series adapt, not as predicting which datasets improve. Penalising the filter for drifting away from a proper wavelet also leaves MSE unchanged (Appendix~\ref{app:vmreg}), so its exact shape is not what drives accuracy.

\begin{table}[h!]
\centering
\caption{Frequency content by sampling rate.}
\label{tab:freq}
\begin{tabular}{lccc}
\toprule
Dataset & Power below 2\,h & Nyquist (cyc/h) & Learnable helps? \\ \midrule
Weather  & 10.0\% & 3.0   & yes \\
ETTm1    & 1.9\%  & 2.0   & yes \\
ETTm2    & 0.9\%  & 2.0   & yes \\
ETTh1    & 0.0\%  & 0.5   & no  \\
ETTh2    & 0.0\%  & 0.5   & no  \\
ECL      & 0.0\%  & 0.5   & yes \\
Traffic  & 0.0\%  & 0.5   & no  \\
Exchange & 0.0\%  & 0.02  & no  \\
ILI      & 0.0\%  & 0.003 & no  \\
\bottomrule
\end{tabular}
\end{table}

These measurements support a narrow claim. The filter gets a real share of the gradient and it does move, mainly in the high pass. But how far it moves, and how much it reshapes the decomposition, say nothing about whether the forecast improves. ECL reshapes almost the least of any dataset and gains the most, while ETTh2 reshapes as much as the datasets that benefit and gains nothing. The one path by which the filter could change the patching has at most a small effect (Table~\ref{tab:hybrid_off}). Where learning the filter helps, it does so by changing the decomposition, not by changing where the patches are cut. Variable-length patches themselves still help (§5.3). What adds nothing is letting the learned filter move the cuts.

\subsection{Discussion and reframing}

The three controlled comparisons support a single conclusion. The multi-resolution decomposition is the component that produces the gain. It beats PatchTST even with the filter frozen (§5.1), and the variable-length patches it induces match or beat a tuned uniform grid built over the same decomposition (§5.3). Learning the filter is a conditional effect, from $-5.6\%$ on ECL and $-3.9\%$ on Weather down to no measurable change on five of the nine datasets and a $2.7\%$ loss on Traffic, and it is bounded by an already near-optimal wavelet prior (§5.2, §5.6). Recomputing the boundaries every step is redundant. The cuts keep moving, but moving them does not change the forecast, and the gradient path through the patching has at most a small effect (§5.6). In practice, keep a fixed wavelet, drop the per-step recomputation and treat learning the filter as a per-dataset decision rather than a default. 

Our results qualify the learnable-wavelet literature rather than an addition to it. Prior ablations do not separate learning the filter from using the decomposition (§2). Doing so across a benchmark panel, we find the decomposition structure accounts for most of the improvement and the filter helps only conditionally. Fine sampling produces the gain on two of the three finely sampled series, and on all three the high-pass filter reshapes more than the low-pass. It is not required though, since hourly ECL shows the largest gain of all and we cannot account for it (§5.4, §5.6). Data volume only sets how far the filter drifts and coarsening the sampling leaves the drift unchanged (§5.4). The common assumption is that a learnable wavelet helps broadly. We find it helps only in a narrow case.

\section{Conclusion and Future Work}

We built a forecasting model that learns a DWT filter and drives variable-length patch boundaries from its per-scale energy. We used it to separate three usually conflated choices, the multi-resolution decomposition, the choice of wavelet and training the filter. The recurring finding is that the decomposition structure and the downstream model produce the gain. Learning the filter helps on three of the nine datasets, most on hourly Electricity, and makes Traffic worse and how far the filter drifts does not predict where it helps. Its benefit is bounded because the fixed decomposition already captures most of the gain. A missing gradient is not the cause, since the filter receives gradient and moves but the forecast barely changes. The variable-length partition should be kept, since it uses 48\% to 63\% fewer tokens than a grid at the same width and a tuned uniform grid beats it on only 9 of 36 settings (§5.3). Its value lies in the patch count and the band tags the decomposition sets rather than in the exact cut positions, since random cuts at the same count and tags forecast almost as well as ours.

\paragraph{Limitations.} The main open item is a rerun at a single model size. Model capacity is not held constant across the mechanism runs and we have not yet measured how much of the reported filter effect that accounts for. The Electricity gain itself is unexplained. It is not separated from the hourly datasets that do not gain by anything we have measured so far. The baselines run at their published configurations, which were not tuned under our budget, so the tie with TimeXer in Table~\ref{tab:main} compares published settings, not equal tuning effort.

\paragraph{Future work.} (1) A boundary-aware model. Our cuts follow one fixed rule, band energy, and random cuts at the same count and tags forecast almost as well (§5.3). We cannot yet tell whether placement is unimportant or whether energy is a poor signal for it, because attention across patches can rejoin whatever a cut splits. The next step is a model in which the dependencies between patches follow the boundaries, with cuts aligned to regime changes and attention that is aware of them. That is where cut placement should start to matter. (2) Scale-gating, masking low-energy scales to stop over-decomposition weakening cross-attention. (3) Timeseries for language models. The same learnable multi-resolution tokeniser could run in the other direction, as a front end that lets language and foundation models read time series in a structured way. This is the counterpart to the byte-tokenisation ideas (BLT~\citep{pagnoni2024blt}, From Bytes to Ideas~\citep{videau2025bytes}) that inspired it (§1.1).

\newpage
\bibliography{litreview_refs}
\bibliographystyle{iclr2025_conference}

\newpage
\appendix

\section{Datasets and splits}\label{app:datasets}

Table~\ref{tab:datasets} lists the nine benchmarks with their splits, channel counts and sampling rates.

\begin{table}[h!]\centering
\begin{threeparttable}
\caption{Benchmark statistics. Lookback $L{=}96$ (ILI $36$); horizons $\{96,192,336,720\}$ (ILI $\{24,36,48,60\}$). Splits follow the Time-Series-Library convention.}
\label{tab:datasets}
\begin{tabular*}{\textwidth}{@{\extracolsep{\fill}}lccccc@{}}
\toprule
Dataset & Channels & Frequency & Timesteps & Split (train/val/test) & Seed $\sigma$ (MSE) \\
\midrule
ETTh1 & 7 & 1 hour & 17{,}420 & 12/4/4 months & 0.0034 \\
ETTh2 & 7 & 1 hour & 17{,}420 & 12/4/4 months & 0.0037 \\
ETTm1 & 7 & 15 min & 69{,}680 & 12/4/4 months & 0.0006 \\
ETTm2 & 7 & 15 min & 69{,}680 & 12/4/4 months & 0.0028 \\
Weather & 21 & 10 min & 52{,}696 & 70/10/20\% & 0.0005 \\
Exchange & 8 & 1 day & 7{,}587 & 70/10/20\% & 0.0129 \\
ILI & 7 & 1 week & 966 & 70/10/20\% & 0.0211 \\
ECL & 321 & 1 hour & 26{,}304 & 70/10/20\% & 0.0001 \\
Traffic & 862 & 1 hour & 17{,}544 & 70/10/20\% & 0.0007 \\
\bottomrule
\end{tabular*}
\begin{tablenotes}[flushleft]\footnotesize
\item Seed $\sigma$ is the standard deviation of the four-horizon mean MSE over seeds $\{2021,42,123\}$.
\end{tablenotes}
\end{threeparttable}
\end{table}

\begin{table}[h]
\caption{Seed standard deviation of the four-horizon mean MSE over seeds $\{2021, 42, 123\}$ for the models in Table~\ref{tab:dynpatch}. A dash marks a cell with no complete three-seed result.}
\label{tab:dynpatch_sd}
\centering\small
\begin{tabular}{lccccc}
\toprule
Dataset & Ours & EntroPE & Pathformer & TimeMosaic & HDMixer \\
\midrule
ETTh1 & 0.0034 & 0.0015 & 0.0008 & 0.0056 & 0.0030 \\
ETTh2 & 0.0037 & 0.0019 & 0.0039 & 0.0019 & 0.0026 \\
ETTm1 & 0.0006 & 0.0024 & 0.0027 & 0.0010 & -- \\
ETTm2 & 0.0028 & 0.0014 & 0.0007 & 0.0011 & -- \\
Weather & 0.0005 & 0.0006 & -- & 0.0002 & -- \\
ECL & 0.0001 & 0.0007 & 0.0011 & 0.0010 & -- \\
Traffic & 0.0007 & 0.0051 & -- & 0.0001 & -- \\
\bottomrule
\end{tabular}
\end{table}

\newpage
\section{Per-task hyperparameters}\label{app:hparams}

Table~\ref{tab:hparams} gives the Optuna-selected architecture for every task.

\begin{table}[h!]\centering\small\setlength{\tabcolsep}{4pt}
\caption{Optuna-selected per-task hyperparameters (lookback $L{=}96$, ILI $36$). $H$ is the forecast horizon, $d$ the model width, $d_{ff}$ the FFN width, $L_e$ the encoder layers and $n_h$ the attention heads. The three dropout rates are the main, fully-connected and head rates. $b$ is the base-patch width and FW the filter width in taps, the number of coefficients in $h$.}
\label{tab:hparams}
\begin{tabular}{llccccccccccc}
\toprule
Data & $H$ & $d$ & $d_{ff}$ & $L_e$ & $n_h$ & drop & fc & head & wavelet & level & $b$ & FW \\
\midrule
ETTh1 & 96 & 512 & 2048 & 4 & 16 & 0.2 & 0.1 & 0.1 & db8 & 3 & 8 & 16 \\
ETTh1 & 192 & 512 & 2048 & 4 & 2 & 0.3 & 0.3 & 0.2 & db8 & 2 & 16 & 16 \\
ETTh1 & 336 & 512 & 2048 & 4 & 16 & 0.3 & 0.1 & 0.2 & db8 & 2 & 16 & 16 \\
ETTh1 & 720 & 256 & 1024 & 4 & 8 & 0.0 & 0.3 & 0.2 & db8 & 4 & 8 & 16 \\
\midrule
ETTh2 & 96 & 512 & 512 & 1 & 16 & 0.0 & 0.2 & 0.2 & db8 & 6 & 2 & 16 \\
ETTh2 & 192 & 512 & 512 & 1 & 16 & 0.3 & 0.2 & 0.2 & db8 & 6 & 4 & 16 \\
ETTh2 & 336 & 512 & 512 & 2 & 4 & 0.2 & 0.0 & 0.2 & db8 & 6 & 4 & 16 \\
ETTh2 & 720 & 512 & 512 & 1 & 2 & 0.2 & 0.2 & 0.3 & db8 & 6 & 8 & 16 \\
\midrule
ETTm1 & 96 & 256 & 2048 & 3 & 2 & 0.1 & 0.0 & 0.0 & db8 & 3 & 2 & 16 \\
ETTm1 & 192 & 512 & 256 & 4 & 2 & 0.2 & 0.1 & 0.1 & db8 & 4 & 8 & 16 \\
ETTm1 & 336 & 256 & 2048 & 4 & 2 & 0.1 & 0.0 & 0.0 & db8 & 4 & 16 & 16 \\
ETTm1 & 720 & 512 & 2048 & 4 & 4 & 0.3 & 0.3 & 0.1 & db8 & 3 & 2 & 16 \\
\midrule
ETTm2 & 96 & 128 & 1024 & 1 & 16 & 0.0 & 0.2 & 0.1 & db8 & 6 & 2 & 16 \\
ETTm2 & 192 & 128 & 256 & 3 & 8 & 0.3 & 0.2 & 0.2 & db8 & 5 & 2 & 16 \\
ETTm2 & 336 & 64 & 256 & 2 & 16 & 0.0 & 0.3 & 0.1 & db8 & 6 & 2 & 16 \\
ETTm2 & 720 & 128 & 256 & 3 & 8 & 0.0 & 0.0 & 0.2 & db8 & 6 & 2 & 16 \\
\midrule
Weather & 96 & 256 & 2048 & 4 & 16 & 0.0 & 0.3 & 0.1 & db8 & 6 & 8 & 16 \\
Weather & 192 & 256 & 512 & 4 & 8 & 0.0 & 0.3 & 0.0 & db8 & 6 & 8 & 16 \\
Weather & 336 & 512 & 256 & 1 & 8 & 0.2 & 0.2 & 0.1 & db8 & 6 & 4 & 16 \\
Weather & 720 & 512 & 256 & 2 & 2 & 0.0 & 0.1 & 0.1 & db8 & 6 & 8 & 16 \\
\midrule
Exchange & 96 & 512 & 512 & 4 & 8 & 0.3 & 0.0 & 0.0 & haar & 2 & 16 & 2 \\
Exchange & 192 & 512 & 512 & 4 & 4 & 0.2 & 0.1 & 0.1 & db8 & 5 & 8 & 16 \\
Exchange & 336 & 512 & 512 & 1 & 8 & 0.3 & 0.3 & 0.0 & haar & 3 & 2 & 2 \\
Exchange & 720 & 512 & 256 & 1 & 16 & 0.2 & 0.1 & 0.1 & db8 & 4 & 2 & 16 \\
\midrule
ILI & 24 & 512 & 1024 & 4 & 4 & 0.3 & 0.3 & 0.0 & db8 & 6 & 2 & 16 \\
ILI & 36 & 512 & 2048 & 4 & 2 & 0.3 & 0.0 & 0.0 & db8 & 6 & 4 & 16 \\
ILI & 48 & 512 & 2048 & 4 & 16 & 0.3 & 0.1 & 0.0 & db8 & 6 & 2 & 16 \\
ILI & 60 & 512 & 2048 & 4 & 8 & 0.3 & 0.0 & 0.0 & db8 & 6 & 2 & 16 \\
\midrule
ECL & 96 & 512 & 2048 & 4 & 8 & 0.0 & 0.3 & 0.0 & db8 & 3 & 8 & 16 \\
ECL & 192 & 512 & 2048 & 4 & 8 & 0.0 & 0.3 & 0.0 & db8 & 4 & 2 & 16 \\
ECL & 336 & 512 & 2048 & 4 & 8 & 0.0 & 0.1 & 0.0 & db8 & 4 & 16 & 16 \\
ECL & 720 & 512 & 2048 & 4 & 16 & 0.0 & 0.3 & 0.0 & db8 & 5 & 2 & 16 \\
\midrule
Traffic & 96 & 512 & 2048 & 4 & 8 & 0.1 & 0.2 & 0.0 & db8 & 6 & 2 & 16 \\
Traffic & 192 & 512 & 512 & 1 & 16 & 0.3 & 0.0 & 0.0 & db8 & 6 & 2 & 16 \\
Traffic & 336 & 512 & 256 & 4 & 16 & 0.2 & 0.1 & 0.0 & db8 & 6 & 4 & 16 \\
Traffic & 720 & 512 & 2048 & 1 & 8 & 0.3 & 0.1 & 0.0 & db8 & 5 & 2 & 16 \\
\bottomrule
\end{tabular}
\end{table}

\newpage
\section{Full per-horizon results}\label{app:perhorizon}

Table~\ref{tab:perhorizon} gives every dataset and horizon cell.

\begin{table}[h!]\centering\scriptsize\setlength{\tabcolsep}{2.5pt}
\caption{Per-horizon test MSE/MAE at lookback $L{=}96$ (ILI $L{=}36$). Best per row per metric in \textbf{bold}. Ours is the 3-seed mean $\{2021,42,123\}$.}
\label{tab:perhorizon}
\resizebox{\textwidth}{!}{%
\begin{tabular}{l c *{18}{c}}
\toprule
& & \multicolumn{2}{c}{Ours} & \multicolumn{2}{c}{Autoformer} & \multicolumn{2}{c}{PatchTST} & \multicolumn{2}{c}{TimesNet} & \multicolumn{2}{c}{Transformer} & \multicolumn{2}{c}{TimeMixer} & \multicolumn{2}{c}{TimeXer} & \multicolumn{2}{c}{iTransformer} & \multicolumn{2}{c}{DLinear} \\
Data & $H$ & MSE & MAE & MSE & MAE & MSE & MAE & MSE & MAE & MSE & MAE & MSE & MAE & MSE & MAE & MSE & MAE & MSE & MAE \\
\midrule
\multirow{4}{*}{ETTh1} & 96 & \textbf{0.368} & \textbf{0.397} & 0.479 & 0.476 & \underline{0.379} & 0.399 & 0.389 & 0.412 & 0.862 & 0.726 & 0.381 & \underline{0.398} & 0.382 & 0.403 & 0.386 & 0.405 & 0.396 & 0.411 \\
 & 192 & \textbf{0.408} & \textbf{0.427} & 0.525 & 0.492 & \underline{0.425} & 0.431 & 0.439 & 0.442 & 0.945 & 0.771 & 0.442 & \underline{0.430} & 0.429 & 0.435 & 0.441 & 0.436 & 0.445 & 0.440 \\
 & 336 & \textbf{0.443} & \textbf{0.448} & 0.509 & 0.492 & 0.469 & 0.457 & 0.495 & 0.471 & 1.017 & 0.826 & 0.501 & 0.460 & \underline{0.467} & \underline{0.449} & 0.491 & 0.461 & 0.487 & 0.465 \\
 & 720 & \textbf{0.465} & \textbf{0.479} & 0.545 & 0.521 & 0.523 & 0.507 & 0.518 & 0.494 & 0.958 & 0.771 & 0.544 & 0.505 & \underline{0.499} & \underline{0.493} & 0.509 & 0.494 & 0.513 & 0.510 \\
\midrule
\multirow{4}{*}{ETTh2} & 96 & \underline{0.287} & 0.342 & 0.368 & 0.406 & 0.294 & 0.347 & 0.337 & 0.371 & 2.583 & 1.289 & \underline{0.287} & \underline{0.340} & \textbf{0.285} & \textbf{0.338} & 0.300 & 0.350 & 0.341 & 0.395 \\
 & 192 & \underline{0.365} & \underline{0.392} & 0.463 & 0.462 & 0.372 & 0.397 & 0.405 & 0.415 & 5.882 & 1.982 & 0.391 & 0.405 & \textbf{0.363} & \textbf{0.389} & 0.382 & 0.400 & 0.482 & 0.479 \\
 & 336 & \textbf{0.410} & \underline{0.425} & 0.481 & 0.482 & 0.426 & 0.440 & 0.454 & 0.451 & 6.440 & 2.075 & 0.427 & 0.431 & \underline{0.414} & \textbf{0.424} & 0.424 & 0.432 & 0.593 & 0.542 \\
 & 720 & \textbf{0.424} & \textbf{0.444} & 0.488 & 0.495 & 0.435 & 0.456 & 0.434 & 0.448 & 3.183 & 1.438 & 0.468 & 0.468 & 0.448 & 0.453 & \underline{0.426} & \underline{0.445} & 0.840 & 0.661 \\
\midrule
\multirow{4}{*}{ETTm1} & 96 & \textbf{0.315} & \underline{0.357} & 0.567 & 0.507 & 0.327 & 0.366 & 0.334 & 0.375 & 0.726 & 0.620 & \underline{0.318} & 0.358 & \underline{0.318} & \textbf{0.356} & 0.341 & 0.376 & 0.346 & 0.374 \\
 & 192 & \textbf{0.355} & \textbf{0.380} & 0.541 & 0.497 & 0.368 & 0.390 & 0.408 & 0.414 & 0.863 & 0.712 & 0.364 & 0.386 & \underline{0.362} & \underline{0.383} & 0.381 & 0.395 & 0.382 & 0.391 \\
 & 336 & \textbf{0.385} & \underline{0.404} & 0.615 & 0.528 & 0.399 & 0.408 & 0.416 & 0.422 & 1.007 & 0.767 & \underline{0.389} & \textbf{0.403} & 0.395 & 0.407 & 0.419 & 0.419 & 0.415 & 0.415 \\
 & 720 & \textbf{0.441} & \textbf{0.438} & 0.589 & 0.528 & 0.458 & 0.445 & 0.486 & 0.461 & 1.119 & 0.823 & \underline{0.451} & 0.442 & 0.452 & \underline{0.441} & 0.486 & 0.456 & 0.473 & 0.451 \\
\midrule
\multirow{4}{*}{ETTm2} & 96 & \underline{0.174} & 0.260 & 0.230 & 0.313 & 0.184 & 0.269 & 0.189 & 0.266 & 0.512 & 0.521 & 0.176 & \underline{0.258} & \textbf{0.171} & \textbf{0.256} & 0.184 & 0.267 & 0.193 & 0.293 \\
 & 192 & 0.242 & \underline{0.306} & 0.281 & 0.340 & 0.247 & \underline{0.306} & 0.252 & 0.307 & 1.632 & 0.939 & \underline{0.238} & \textbf{0.299} & \textbf{0.236} & \textbf{0.299} & 0.253 & 0.312 & 0.285 & 0.361 \\
 & 336 & 0.302 & 0.342 & 0.339 & 0.374 & 0.310 & 0.349 & 0.321 & 0.349 & 1.115 & 0.798 & \underline{0.299} & \underline{0.339} & \textbf{0.295} & \textbf{0.338} & 0.315 & 0.352 & 0.385 & 0.429 \\
 & 720 & 0.403 & 0.403 & 0.481 & 0.454 & 0.421 & 0.415 & 0.418 & 0.405 & 2.952 & 1.297 & \underline{0.393} & \underline{0.397} & \textbf{0.392} & \textbf{0.393} & 0.412 & 0.406 & 0.556 & 0.523 \\
\midrule
\multirow{4}{*}{Weather} & 96 & 0.164 & 0.210 & 0.314 & 0.365 & 0.178 & 0.221 & 0.169 & 0.219 & 0.375 & 0.409 & \underline{0.162} & \underline{0.209} & \textbf{0.157} & \textbf{0.205} & 0.175 & 0.216 & 0.196 & 0.256 \\
 & 192 & 0.210 & 0.253 & 0.326 & 0.377 & 0.221 & 0.256 & 0.225 & 0.265 & 0.549 & 0.525 & \underline{0.207} & \underline{0.251} & \textbf{0.204} & \textbf{0.247} & 0.225 & 0.258 & 0.239 & 0.299 \\
 & 336 & 0.265 & 0.293 & 0.351 & 0.386 & 0.280 & 0.298 & 0.281 & 0.304 & 0.633 & 0.583 & \underline{0.263} & \underline{0.292} & \textbf{0.260} & \textbf{0.290} & 0.280 & 0.298 & 0.281 & 0.331 \\
 & 720 & \textbf{0.341} & \textbf{0.342} & 0.440 & 0.441 & 0.358 & 0.350 & 0.359 & 0.354 & 0.944 & 0.711 & \underline{0.344} & \underline{0.346} & \textbf{0.341} & \textbf{0.342} & 0.361 & 0.351 & 0.345 & 0.382 \\
\midrule
\multirow{4}{*}{Exchange} & 96 & 0.088 & 0.208 & 0.143 & 0.274 & 0.095 & 0.213 & 0.105 & 0.235 & 0.543 & 0.577 & \textbf{0.086} & \textbf{0.204} & 0.091 & 0.209 & \underline{0.087} & \underline{0.207} & 0.098 & 0.233 \\
 & 192 & \underline{0.184} & \underline{0.305} & 0.291 & 0.391 & \underline{0.184} & \underline{0.305} & 0.225 & 0.345 & 0.973 & 0.753 & 0.192 & 0.311 & 0.190 & 0.309 & \textbf{0.180} & \textbf{0.303} & 0.186 & 0.325 \\
 & 336 & \underline{0.332} & \textbf{0.418} & 0.467 & 0.510 & 0.353 & 0.433 & 0.363 & 0.439 & 1.476 & 0.947 & 0.359 & 0.436 & 0.369 & 0.437 & 0.334 & \underline{0.420} & \textbf{0.327} & 0.435 \\
 & 720 & 0.871 & 0.702 & 1.118 & 0.822 & 0.952 & 0.733 & 0.931 & 0.735 & 2.199 & 1.189 & 0.916 & 0.721 & 0.899 & 0.714 & \underline{0.856} & \underline{0.700} & \textbf{0.749} & \textbf{0.664} \\
\midrule
\multirow{4}{*}{ILI} & 24 & \underline{1.880} & 0.915 & 3.304 & 1.273 & 2.267 & \underline{0.889} & \textbf{1.790} & \textbf{0.885} & 4.571 & 1.405 & 2.538 & 1.002 & 2.644 & 1.022 & 3.153 & 1.210 & 5.061 & 1.710 \\
 & 36 & \textbf{1.947} & \textbf{0.908} & 3.377 & 1.257 & \underline{2.318} & \underline{0.909} & 2.633 & 0.981 & 4.522 & 1.442 & 2.718 & 1.086 & 2.492 & 0.981 & 2.893 & 1.144 & 4.414 & 1.550 \\
 & 48 & \textbf{1.977} & \underline{0.915} & 3.077 & 1.179 & \underline{2.056} & \textbf{0.877} & 2.577 & 0.937 & 4.799 & 1.478 & 2.505 & 1.011 & 2.526 & 0.991 & 2.870 & 1.148 & 4.109 & 1.473 \\
 & 60 & 2.192 & 0.979 & 2.817 & 1.131 & \underline{2.021} & \underline{0.921} & \textbf{1.965} & \textbf{0.891} & 5.083 & 1.522 & 2.564 & 1.025 & 2.646 & 1.039 & 2.939 & 1.168 & 4.234 & 1.482 \\
\midrule
\multirow{4}{*}{ECL} & 96 & 0.159 & 0.250 & 0.204 & 0.320 & 0.180 & 0.273 & 0.168 & 0.272 & 0.333 & 0.423 & 0.156 & 0.247 & \textbf{0.140} & \underline{0.242} & \underline{0.148} & \textbf{0.240} & 0.210 & 0.302 \\
 & 192 & 0.171 & \underline{0.260} & 0.222 & 0.330 & 0.187 & 0.280 & 0.188 & 0.291 & 0.371 & 0.447 & 0.170 & \underline{0.260} & \textbf{0.157} & \textbf{0.256} & \underline{0.164} & \textbf{0.256} & 0.210 & 0.305 \\
 & 336 & 0.188 & 0.278 & 0.306 & 0.382 & 0.204 & 0.296 & 0.206 & 0.306 & 0.473 & 0.509 & 0.187 & 0.278 & \textbf{0.176} & \underline{0.274} & \underline{0.178} & \textbf{0.271} & 0.223 & 0.319 \\
 & 720 & 0.227 & \underline{0.311} & 0.406 & 0.437 & 0.246 & 0.328 & \underline{0.224} & 0.320 & 0.460 & 0.511 & 0.228 & 0.312 & \textbf{0.211} & \textbf{0.307} & 0.228 & 0.312 & 0.258 & 0.350 \\
\midrule
\multirow{4}{*}{Traffic} & 96 & 0.475 & 0.297 & 0.661 & 0.398 & 0.459 & 0.298 & 0.589 & 0.315 & 0.663 & 0.368 & 0.468 & 0.293 & \underline{0.429} & \underline{0.272} & \textbf{0.393} & \textbf{0.268} & 0.697 & 0.429 \\
 & 192 & 0.486 & 0.315 & 0.670 & 0.429 & 0.469 & 0.301 & 0.615 & 0.325 & 0.656 & 0.361 & 0.509 & 0.302 & \underline{0.446} & \underline{0.281} & \textbf{0.413} & \textbf{0.277} & 0.647 & 0.407 \\
 & 336 & 0.506 & 0.312 & 0.683 & 0.426 & 0.483 & 0.307 & 0.638 & 0.340 & 0.656 & 0.359 & 0.521 & 0.308 & \underline{0.470} & \underline{0.289} & \textbf{0.425} & \textbf{0.283} & 0.653 & 0.410 \\
 & 720 & 0.537 & 0.337 & 0.655 & 0.411 & \underline{0.518} & 0.326 & 0.662 & 0.350 & 0.693 & 0.376 & 0.552 & 0.324 & 0.519 & \underline{0.307} & \textbf{0.460} & \textbf{0.301} & 0.694 & 0.429 \\
\bottomrule
\end{tabular}}
\end{table}

\begin{table*}[h!]
\caption{Per-horizon test MSE/MAE of the dynamic-patching methods at lookback $L{=}96$. Best per row per metric in \textbf{bold}, second best \underline{underlined}. Every model is the mean over seeds $\{2021, 42, 123\}$. HDMixer releases configurations only for ETTh1 and ETTh2. Pathformer diverged on Weather and its Traffic runs did not complete. All other runs converged.}
\label{tab:dynpatch_h}
\centering
\resizebox{\textwidth}{!}{%
\begin{tabular}{ll cc cc cc cc cc}
\toprule
 & & \multicolumn{2}{c}{Ours} & \multicolumn{2}{c}{EntroPE} & \multicolumn{2}{c}{Pathformer} & \multicolumn{2}{c}{TimeMosaic} & \multicolumn{2}{c}{HDMixer} \\
\cmidrule(lr){3-4}\cmidrule(lr){5-6}\cmidrule(lr){7-8}\cmidrule(lr){9-10}\cmidrule(lr){11-12}
Data & $H$ & MSE & MAE & MSE & MAE & MSE & MAE & MSE & MAE & MSE & MAE \\
\midrule
ETTh1 & 96 & \textbf{0.368} & 0.397 & 0.390 & 0.406 & 0.385 & \underline{0.390} & \underline{0.369} & \textbf{0.385} & 0.387 & 0.401 \\
 & 192 & \textbf{0.408} & 0.427 & 0.432 & 0.430 & 0.444 & \underline{0.421} & \underline{0.428} & \textbf{0.418} & 0.441 & 0.428 \\
 & 336 & \textbf{0.443} & 0.448 & 0.469 & 0.447 & 0.462 & \textbf{0.429} & 0.512 & 0.469 & \underline{0.449} & \underline{0.432} \\
 & 720 & \underline{0.465} & 0.479 & \textbf{0.463} & \underline{0.466} & 0.490 & \textbf{0.463} & 0.492 & 0.470 & 0.486 & 0.470 \\
\midrule
ETTh2 & 96 & 0.287 & 0.342 & 0.305 & 0.353 & \underline{0.283} & \textbf{0.334} & 0.302 & 0.343 & \textbf{0.280} & \underline{0.336} \\
 & 192 & 0.365 & 0.392 & 0.395 & 0.411 & \textbf{0.351} & \textbf{0.377} & 0.365 & \underline{0.383} & \underline{0.361} & 0.387 \\
 & 336 & 0.410 & 0.425 & 0.405 & 0.430 & \textbf{0.341} & \textbf{0.378} & 0.408 & 0.417 & \underline{0.347} & \underline{0.386} \\
 & 720 & 0.424 & 0.444 & 0.418 & 0.443 & \textbf{0.409} & \textbf{0.429} & \underline{0.415} & \underline{0.433} & 0.422 & 0.440 \\
\midrule
ETTm1 & 96 & \underline{0.315} & 0.357 & 0.325 & 0.359 & \underline{0.315} & \underline{0.345} & \textbf{0.309} & \textbf{0.337} & -- & -- \\
 & 192 & \textbf{0.355} & 0.380 & 0.373 & 0.386 & 0.364 & \underline{0.369} & \underline{0.362} & \textbf{0.367} & -- & -- \\
 & 336 & \textbf{0.385} & \underline{0.404} & 0.394 & 0.405 & \underline{0.388} & \textbf{0.390} & 0.396 & \textbf{0.390} & -- & -- \\
 & 720 & \textbf{0.441} & 0.438 & \underline{0.453} & 0.436 & 0.458 & \textbf{0.427} & 0.462 & \underline{0.429} & -- & -- \\
\midrule
ETTm2 & 96 & 0.174 & 0.260 & 0.186 & 0.270 & \textbf{0.170} & \underline{0.250} & \underline{0.171} & \textbf{0.247} & -- & -- \\
 & 192 & 0.242 & \underline{0.306} & 0.254 & 0.313 & \textbf{0.235} & \textbf{0.293} & \underline{0.238} & \textbf{0.293} & -- & -- \\
 & 336 & \underline{0.302} & 0.342 & 0.315 & 0.352 & \textbf{0.300} & \textbf{0.335} & 0.305 & \underline{0.336} & -- & -- \\
 & 720 & 0.403 & 0.403 & 0.423 & 0.417 & \textbf{0.388} & \textbf{0.388} & \underline{0.398} & \underline{0.391} & -- & -- \\
\midrule
Weather & 96 & \textbf{0.164} & \underline{0.210} & 0.171 & 0.219 & -- & -- & \underline{0.166} & \textbf{0.198} & -- & -- \\
 & 192 & \textbf{0.210} & \underline{0.253} & \underline{0.215} & 0.256 & -- & -- & 0.216 & \textbf{0.244} & -- & -- \\
 & 336 & \textbf{0.265} & \underline{0.293} & \underline{0.271} & 0.297 & -- & -- & 0.273 & \textbf{0.287} & -- & -- \\
 & 720 & \textbf{0.341} & \underline{0.342} & \underline{0.342} & 0.343 & -- & -- & 0.352 & \textbf{0.339} & -- & -- \\
\midrule
ECL & 96 & \underline{0.159} & \underline{0.250} & 0.172 & 0.267 & \textbf{0.142} & \textbf{0.233} & 0.166 & 0.251 & -- & -- \\
 & 192 & \underline{0.171} & \underline{0.260} & 0.182 & 0.275 & \textbf{0.162} & \textbf{0.251} & 0.178 & 0.262 & -- & -- \\
 & 336 & \underline{0.188} & 0.278 & 0.196 & 0.288 & \textbf{0.175} & \textbf{0.265} & 0.192 & \underline{0.277} & -- & -- \\
 & 720 & 0.227 & 0.311 & 0.236 & 0.324 & \textbf{0.201} & \textbf{0.291} & \underline{0.224} & \underline{0.303} & -- & -- \\
\midrule
Traffic & 96 & \underline{0.475} & \underline{0.297} & 0.504 & 0.300 & -- & -- & \textbf{0.423} & \textbf{0.266} & -- & -- \\
 & 192 & \underline{0.486} & 0.315 & 0.504 & \underline{0.297} & -- & -- & \textbf{0.443} & \textbf{0.273} & -- & -- \\
 & 336 & 0.506 & 0.312 & \underline{0.504} & \underline{0.300} & -- & -- & \textbf{0.459} & \textbf{0.278} & -- & -- \\
 & 720 & \underline{0.537} & 0.337 & 0.543 & \underline{0.316} & -- & -- & \textbf{0.491} & \textbf{0.295} & -- & -- \\
\bottomrule
\end{tabular}}
\end{table*}

\newpage
\section{Patch counts and lengths}\label{app:patchcounts}
\begin{figure}[h!]
  \centering
  \includegraphics[width=\linewidth]{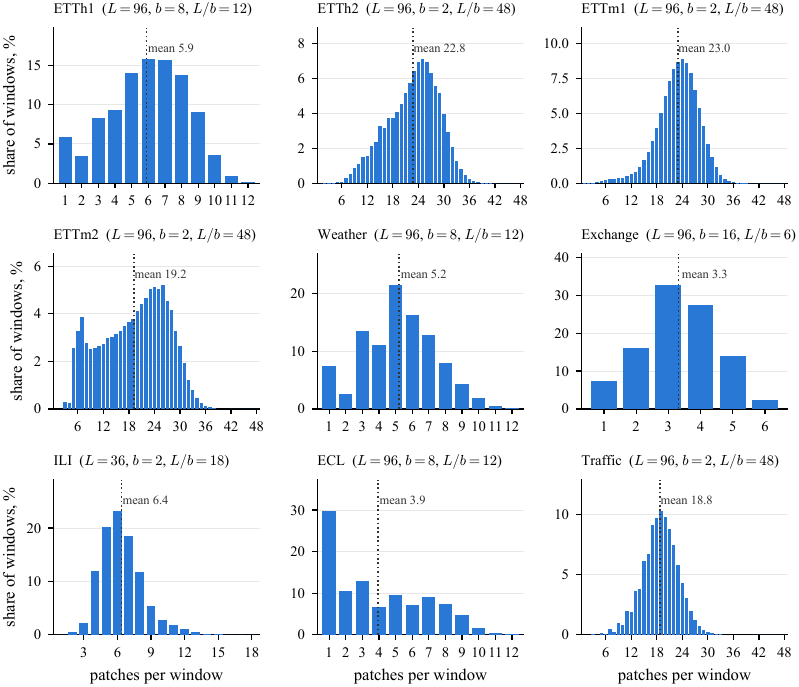}
  \caption{Patches per window and channel on the training split at the 96-step horizon (ILI 24), one panel per dataset at its tuned base window width. The dotted line is the mean. A uniform grid at the same width would use $L/b$ patches, the right end of each axis.}
  \label{fig:patchcounts}
\end{figure}

Table~\ref{tab:patchcounts} counts the patches the merge produces per window and channel on the training split, at each task's tuned base window width $b$. The boundaries come from the frozen db8 filters, so the counts are the same for every mode and seed. A uniform grid at the same width would use $L/b$ tokens. Merging uses 48\% to 63\% fewer tokens, see the Tokens saved column. Figure~\ref{fig:patchcounts} shows the distribution at the 96-step horizon (ILI 24). The random-merging control of §5.3 keeps these counts and moves 53\% to 73\% of the cuts.

\begin{table}[h!]
\centering\small
\caption{Merged patch counts on the training split, four-horizon mean. $b$ lists the tuned base window widths across the four horizons, $L/b$ is the uniform token count at those widths, tokens saved is $1-\text{patches}/(L/b)$ and the last column is the mean patch length in time steps.}
\label{tab:patchcounts}
\begin{tabular}{lcccccc}
\toprule
Dataset & $L$ & $b$ & $L/b$ & Patches & Tokens saved (\%) & Mean length (steps) \\
\midrule
ETTh1    & 96 & 8, 16    & 9.0  & 4.3  & 53 & 24.2 \\
ETTh2    & 96 & 2, 4, 8  & 27.0 & 14.0 & 48 & 8.4 \\
ETTm1    & 96 & 2, 8, 16 & 28.5 & 13.8 & 52 & 13.3 \\
ETTm2    & 96 & 2        & 48.0 & 19.8 & 59 & 4.9 \\
Weather  & 96 & 4, 8     & 15.0 & 6.0  & 60 & 16.7 \\
Exchange & 96 & 2, 8, 16 & 28.5 & 13.2 & 54 & 13.5 \\
ILI      & 36 & 2, 4     & 15.8 & 5.9  & 62 & 6.2 \\
ECL      & 96 & 2, 8, 16 & 28.5 & 10.6 & 63 & 23.2 \\
Traffic  & 96 & 2, 4     & 42.0 & 16.5 & 61 & 6.3 \\
\bottomrule
\end{tabular}
\end{table}

\newpage
\section{Configuration provenance for the benchmark}\label{app:provenance}

Every result in Tables~\ref{tab:main} and~\ref{tab:dynpatch} is our own re-run in one harness at $L{=}96$ (ILI 36). No number is copied from another paper. Table~\ref{tab:provenance} records where each run's configuration came from.

\begin{table}[h!]
\caption{Configuration source per model and dataset. \textbf{O} is the authors' own configuration for that dataset at our lookback, run unchanged, from the benchmark library or the model's own repository. \textbf{T} means we found no configuration at our lookback in the library, the model's repository or its paper, so we transfer the model's nearest published configuration. \textbf{F} is the authors' released configuration with only the lookback set to 96. A dash means not reported.}
\label{tab:provenance}
\centering\small
\begin{tabular}{lccccccccc}
\toprule
Model & ETTh1 & ETTh2 & ETTm1 & ETTm2 & Weather & Exchange & ILI & ECL & Traffic \\
\midrule
\multicolumn{10}{l}{\textit{Standard baselines}} \\
PatchTST & O & O & O & O & O & O & O & O & O \\
TimesNet & O & O & O & O & O & O & O & O & O \\
Autoformer & O & O & O & O & O & O & O & O & O \\
Transformer & O & O & O & O & O & O & O & O & O \\
TimeXer & O & O & O & O & O & T & T & O & O \\
TimeMixer & O & O & O & O & O & T & T & O & O \\
iTransformer & O & O & O & O & O & O & T & O & O \\
DLinear$^{a}$ & O & T & T & T & T & T & T & O & T \\
\midrule
\multicolumn{10}{l}{\textit{Dynamic-patching methods}} \\
EntroPE & O & O & O & O & O & -- & -- & O & O \\
Pathformer & O & O & O & O & O & -- & -- & O & O \\
TimeMosaic & F & F & F & F & F & -- & -- & O & O \\
HDMixer & F & F & -- & -- & -- & -- & -- & -- & -- \\
\bottomrule
\end{tabular}

\vspace{2pt}
\begin{minipage}{\linewidth}\footnotesize
$^{a}$ DLinear's paper reports results at a lookback of 336, so outside the two library scripts we found no DLinear configuration at 96.
\end{minipage}
\end{table}

\textbf{Where the configurations come from.} The 60 O cells run the authors own scripts, 56 from the benchmark library and 4 from iTransformer's repository. The 12 T cells cover datasets for which we found no configuration at our lookback, so each transfers that model's nearest published configuration, with only the dataset block and lookback changed. The exact scripts for every cell are released with the code.

\textbf{Dynamic-patching methods.} EntroPE and Pathformer are run from their released scripts unchanged. TimeMosaic's released ECL and Traffic scripts already use $L{=}96$. Its ETT and Weather scripts use $L{=}320$, which we set to 96. HDMixer releases scripts only for ETTh1 and ETTh2, at $L{=}336$, which we also set to 96. Pathformer and TimeMosaic fix the random seed inside their training script, so for the three-seed runs we change only that value. EntroPE's paper reports the best of five seeds and its released scripts fix one seed per task. We keep its configuration and run seeds $\{2021, 42, 123\}$, the same seeds as every other model.

\textbf{Does the harness reproduce?} The T cells have no published number to check against, so we check the harness where one exists. TimeXer matches its published ECL and Traffic results exactly, and PatchTST, iTransformer and TimeMixer come within 0.005 MSE. Autoformer, Transformer and DLinear score worse than published, a known seed sensitivity and no conclusion rests on them. If the harness recovers the published numbers where they exist, we can trust it on the cells where they do not.

\section{Significance of Ours vs.\ each baseline}\label{app:significance}
Paired Wilcoxon signed-rank on the nine four-horizon mean MSEs of Table~\ref{tab:main}, our model (Learned) against each baseline. Each dataset counts once, because its four horizons share training data and configuration and are not independent. Table~\ref{tab:significance} reports the test. Our model is significantly better than six of the eight baselines at $p<0.1$ and statistically indistinguishable from iTransformer and TimeXer. With $n=9$ the smallest attainable $p$ is $0.004$, so the test is conservative.

\begin{table}[h!]\centering
\caption{Paired Wilcoxon signed-rank, our model vs.\ each baseline over the nine datasets (four-horizon mean MSE). $\Delta$MSE is ours minus baseline, so negative favours ours.}
\label{tab:significance}
\begin{tabular}{lccc}
\toprule
Comparison & wins/losses & median $\Delta$MSE & Wilcoxon $p$ \\
\midrule
vs.\ Autoformer   & 9/0 & $-0.113$ & $0.004$ \\
vs.\ PatchTST     & 8/1 & $-0.014$ & $0.055$ \\
vs.\ TimesNet     & 9/0 & $-0.037$ & $0.004$ \\
vs.\ Transformer  & 9/0 & $-0.555$ & $0.004$ \\
vs.\ TimeMixer    & 6/3 & $-0.011$ & $0.055$ \\
vs.\ TimeXer      & 5/4 & $-0.007$ & $0.496$ \\
vs.\ iTransformer & 6/3 & $-0.012$ & $0.203$ \\
vs.\ DLinear      & 8/1 & $-0.039$ & $0.012$ \\
\bottomrule
\end{tabular}
\end{table}

\newpage
\section{Implementation, training and reproducibility}\label{app:impl}\label{app:repro}
Our models minimise MSE with Adam (batch size 32, or 16 on ECL and 4 on Traffic, initial learning rate $10^{-4}$, ILI $5\times10^{-4}$) under the \texttt{type1} schedule (halve the rate each epoch), with RevIN instance normalisation on inputs and outputs. Each of our runs trains for up to 10 epochs with early stopping (patience 3) on validation loss. Our model in Table~\ref{tab:main} and the three DWT modes of Table~\ref{tab:mode_ladder} are means over three seeds $\{2021,42,123\}$ and the baselines, Tables~\ref{tab:ladder_seed}, \ref{tab:init_sensitivity} to \ref{tab:hybrid_off} and the Appendix~\ref{app:sweeps} experiments are single-seed \{2021\}. 

Table~\ref{tab:impl} lists the settings held fixed across every task.

\begin{table}[h!]\centering
\caption{Deployed model settings held fixed across tasks. Per-task architecture (\texttt{d\_model}, \texttt{d\_ff}, layers, heads, dropouts, wavelet, level, base-patch, filter width) is selected by Optuna and released with the code.}
\label{tab:impl}
\begin{tabular*}{\textwidth}{@{\extracolsep{\fill}}ll@{}}
\toprule
Setting & Value \\
\midrule
Straight-through strength $\alpha$ & 1.0 \\
Optimiser / schedule & Adam / type1 \\
Batch size & 32 (16 ECL, 4 Traffic) \\
Seeds (main table) & \{2021, 42, 123\} \\
Filter learning rate & same as the model \\
Soft-weight temperature $\tau$ & 1.0 \\
\bottomrule
\end{tabular*}
\end{table}

Datasets are the public benchmarks of Appendix~\ref{app:datasets} with standard Time-Series-Library splits and per-task hyperparameters are in Appendix~\ref{app:hparams}. Code and configurations are available at \url{https://github.com/hafizalsree/dwt-dynamic-patching}.

\section{Sensitivity to filter width, wavelet family and decomposition depth}
\label{app:sweeps}

Three structural axes are varied one at a time with everything else held at the task tuned configuration: wavelet family ($\in \{\text{Haar, db4, db8, db12, db16, db20, sym8, sym12, coif4, coif5}\}$, spanning 2 to 40 taps), decomposition level ($\in \{2, 4, 6, 8\}$) and base-patch size ($b\in \{4,8,16,24\}$). These runs use the main model, so the DWT filter is learnable throughout and each measures how robust the main forecast is to a structural choice. For level and base-patch we report the across-setting coefficient of variation (std/mean of MSE, Table \ref{tab:struct_sens}) and the full curve of MSE against swept quantity (Tables~\ref{tab:level} and~\ref{tab:basepatch}).

\begin{table}[h!]
\centering
\caption{Structural sensitivity (std/mean of MSE across settings, \%). Wavelet over the ten wavelets of Figure~\ref{fig:wav_width}, level over $\{2,4,6,8\}$, base-patch over $\{4,8,16,24\}$.}
\label{tab:struct_sens}
\begin{tabular}{lccc}
\toprule
Dataset & Wavelet family & Level & Base-patch \\
\midrule
ETTh1    & 0.12\%          & 0.23\% & 0.50\% \\
ETTh2    & 1.73\%          & 2.63\% & 1.20\% \\
ETTm1    & 1.31\%          & 0.65\% & 0.74\% \\
ETTm2    & 0.39\%          & 0.67\% & 2.00\% \\
Weather  & 0.38\%          & 0.71\% & 0.10\% \\
Exchange & 1.18\%          & 4.94\% & 2.38\% \\
ILI      & 4.7\%$^{*}$     & 3.41\% & 3.49\% \\
ECL      & 0.79\%          & 0.17\% & 0.10\% \\
Traffic  & 0.78\%          & 0.49\% & 1.27\% \\
\bottomrule
\end{tabular}

{\footnotesize $^{*}$ILI wavelet over the nine non-divergent wavelets; including the db20 divergence (Figure~\ref{fig:wav_width}) it is 21.5\%.}
\end{table}

Within a family the choice reduces to a single axis. Every family we test (Daubechies, Symlets, Coiflets and the Haar special case) is orthogonal at initialisation. The operative difference is filter width, which for order-$N$ Daubechies is $2N$ taps and couples one-to-one with the number of vanishing moments $N$, the count of polynomial degrees the detail filter annihilates. More vanishing moments make the wavelet blind to smoother trend and concentrate genuine fluctuation into sparser detail coefficients, but they require longer support that blurs sharp transients. Width is therefore a single interpretable parameter that trades spectral selectivity against time localisation.

\begin{figure}[h!]
  \centering
  \includegraphics[width=0.85\linewidth]{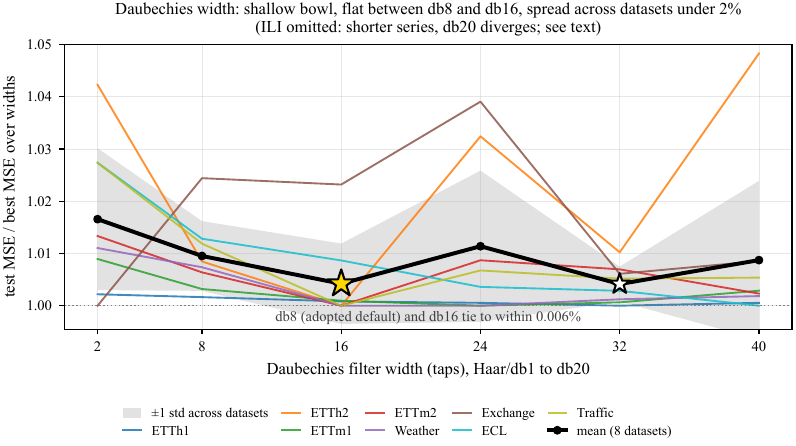}
  \caption{Daubechies filter width, normalised per dataset. Mean bowl (bold) with $\pm1$ std band; shallow, floor flat between db8 and db16. ILI omitted (db20 diverges).}
  \label{fig:wav_width}
\end{figure}

Read as a Daubechies width curve, the result is a shallow bowl rather than a sharp optimum. Averaged across the eight datasets, the MSE spans only about 1.2\% from best to worst width and its floor is at db8 (16 taps). The curvature sits at the extremes. On most datasets the narrowest Haar is the worst width, but two datasets reverse this. On the trend-only Exchange, Haar is instead the best. On ETTh2, the worst is the widest db20. The interior widths from db4 to db16 are nearly flat, so among the usable widths the choice barely matters. Most of these differences fall inside the seed-noise band of §5.1, which makes the bowl a weak preference rather than a strong optimum. Sensitivity is set by the signal and not the dataset size, ETTh1 and ETTh2 share the same training size (8640 steps) yet differ 14-fold in coefficient of variation (0.12\% vs 1.73\%) and identically sampled ETTm1 and ETTm2 differ threefold ($1.31\%$ vs $0.39\%$).

Where MSE does clear the noise it can be the family rather than the width. Coif4 is an upward outlier on ETTm1 (0.327 against 0.313 for the equal-width db12 and sym12) and on the short ILI series db8 beats sym8 at the same 16 taps while db20 diverges outright (a 40 tap filter on a 36-step input), which is why ILI is reported apart from Figure \ref{fig:wav_width}. We therefore adopt db8 (16 taps, eight vanishing moments) as the default on a stronger basis than a modal optimum. It lies inside the seed-noise tie band on all nine datasets so it is never distinguishable from the best available wavelet. Where a dataset has no fast detail at all, such as the daily Exchange, the narrowest Haar is nominally the best, but its whole width curve is within noise, so this is width-insensitivity rather than a Haar optimum. 

\begin{table}[h!]
\centering
\caption{Test MSE by decomposition level (p96; ILI p24). Best per row in \textbf{bold}.}
\label{tab:level}
\begin{tabular}{lcccc}
\toprule
Dataset & 2 & 4 & 6 & 8 \\
\midrule
ETTh1    & 0.3681 & 0.3689 & 0.3691 & \textbf{0.3669} \\
ETTh2    & 0.2934 & 0.2880 & \textbf{0.2837} & 0.3043 \\
ETTm1    & 0.3204 & \textbf{0.3151} & 0.3198 & 0.3187 \\
ETTm2    & 0.1746 & 0.1745 & 0.1724 & \textbf{0.1721} \\
Weather  & 0.1658 & 0.1633 & \textbf{0.1630} & 0.1631 \\
Exchange & \textbf{0.0819} & 0.0912 & 0.0896 & 0.0938 \\
ILI      & 2.1480 & 2.1972 & \textbf{2.0029} & 2.0966 \\
ECL      & \textbf{0.1598} & 0.1600 & 0.1600 & 0.1604 \\
Traffic  & 0.4775 & 0.4785 & \textbf{0.4731} & 0.4758 \\
\bottomrule
\end{tabular}
\end{table}

\begin{table}[h!]
\centering
\caption{Test MSE by base-patch size in time steps (p96; ILI p24). Best per row in \textbf{bold}.}
\label{tab:basepatch}
\begin{tabular}{lcccc}
\toprule
Dataset & 4 & 8 & 16 & 24 \\
\midrule
ETTh1    & 0.3714 & \textbf{0.3662} & 0.3684 & 0.3682 \\
ETTh2    & 0.2907 & \textbf{0.2877} & 0.2971 & 0.2939 \\
ETTm1    & 0.3154 & 0.3149 & 0.3200 & \textbf{0.3139} \\
ETTm2    & \textbf{0.1744} & 0.1756 & 0.1760 & 0.1834 \\
Weather  & 0.1630 & 0.1630 & \textbf{0.1626} & 0.1628 \\
Exchange & 0.0865 & 0.0871 & \textbf{0.0819} & 0.0843 \\
ILI      & 2.1242 & 2.2835 & \textbf{2.0823} & 2.1829 \\
ECL      & 0.1595 & 0.1592 & 0.1595 & \textbf{0.1592} \\
Traffic  & \textbf{0.4796} & 0.4869 & 0.4891 & 0.4945 \\
\bottomrule
\end{tabular}
\end{table}

Decomposition level sets how many frequency bands the transform extracts from the signal, so a dataset's preferred depth reflects how many scales carry real structure. This is clearest on Exchange, an almost pure-trend series whose energy sits in a single coarse band. It is the most level-sensitive dataset (4.9\%, Table~\ref{tab:struct_sens}) and prefers the shallowest level, worsening from 0.082 at the shallowest level to 0.094 because deeper decomposition only splits a near-empty detail region into further noise bands. The other two datasets with real level sensitivity, ETTh2 and ILI prefer a deeper level (6), while the remaining six (ETTh1, ETTm1, ETTm2, Weather, ECL, Traffic) have no dominant scale and are nearly indifferent (level CV under 1\%). Level is therefore data-specific.

Base-patch size is the temporal granularity of the finest tokens, not a count of frequency bands. Its effect across the grid is mild and dataset-specific with no consistent best value (Table \ref{tab:basepatch}) and no signal characteristic separates the choices, ETTm1 and ETTm2 identical in sampling and domain, land opposite ends. The only regularity is that the main Optuna search on a finer grid, settles on the smallest base patch on five of seven datasets. Hence, the fine tokenisation appears mildly helpful.

Because these measurements use the main learnable filter, a low sensitivity can reflect the filter compensating for a suboptimal structural choice rather than genuine indifference. We therefore also ran a frozen-filter arm on ECL and Traffic at matched wavelets. Freezing does not widen the spread across settings, $0.08\%$ and $0.13\%$ frozen against $0.17\%$ and $0.24\%$ learnable, so the insensitivity is a property of the decomposition rather than compensation by the filter. The paired arms also show the §5.1 split cleanly, with every learnable setting beating every frozen one on ECL by about $8\%$ and every frozen setting beating every learnable one on Traffic by about $4.6\%$.

\section{Full mechanism statistics}\label{app:mechanism}

Table~\ref{tab:mechanism} in §5.6 reports the Learned mode only, and drops high-pass drift because it is the product of the two columns beside it. The full set is in Table~\ref{tab:mechanism_full} for both trained, Learned with cached boundaries and Dynamic with boundaries recomputed each step. Drift is each filter's distance from its wavelet initialisation at the final epoch, averaged over the four horizons. Gradient share is the ratio of the gradient norm on the filter to the gradient norm on the rest of the model, at the final epoch. Scale energy shift is the total variation distance between the per-band energy at the first and last epoch, as a percentage.

\begin{table}[h!]\centering\small
\caption{Final-epoch mechanism statistics for both trained modes, all nine
datasets. Both modes were run with the straight-through path off ($\alpha=0$). In Learned mode, $\alpha$ has no effect, so that column is unaffected.}
\label{tab:mechanism_full}
\resizebox{\textwidth}{!}{%
\begin{tabular}{lccccc|ccccc}
\toprule
& \multicolumn{5}{c|}{Learned} & \multicolumn{5}{c}{Dynamic} \\
Dataset & Drift $h$ & Drift $g$ & $g/h$ & Grad share & Scale energy (\%)
        & Drift $h$ & Drift $g$ & $g/h$ & Grad share & Scale energy (\%) \\
\midrule
ETTm2    & 0.070 & 0.099 & 1.43 & 0.78 & 15.9 & 0.072 & 0.090 & 1.25 & 0.90 & 18.1 \\
Weather  & 0.069 & 0.092 & 1.33 & 1.08 & 11.5 & 0.071 & 0.086 & 1.21 & 0.99 & 11.9 \\
ETTh2    & 0.044 & 0.053 & 1.20 & 0.87 & 9.0  & 0.046 & 0.043 & 0.94 & 1.06 & 6.8 \\
ETTm1    & 0.078 & 0.128 & 1.65 & 0.43 & 8.9  & 0.079 & 0.113 & 1.43 & 0.47 & 8.6 \\
ETTh1    & 0.049 & 0.045 & 0.92 & 0.26 & 2.7  & 0.048 & 0.049 & 1.01 & 0.28 & 2.4 \\
ILI      & 0.007 & 0.006 & 0.89 & 0.46 & 2.5  & 0.007 & 0.006 & 0.92 & 0.60 & 2.4 \\
Traffic  & 0.065 & 0.081 & 1.24 & 0.09 & 2.1  & 0.060 & 0.081 & 1.35 & 0.09 & 2.1 \\
ECL      & 0.067 & 0.062 & 0.93 & 0.14 & 1.6  & 0.066 & 0.064 & 0.97 & 0.14 & 2.0 \\
Exchange & 0.019 & 0.023 & 1.26 & 0.57 & 0.8  & 0.020 & 0.024 & 1.20 & 0.57 & 1.2 \\
\bottomrule
\end{tabular}}
\end{table}

Low-pass drift is almost mode-invariant, differing by at most 0.005 on any dataset, while high-pass drift moves by up to 0.015, so the ratio is where the two modes differ. The ordering of gradient share is nearly the same in both, with Weather and ETTh2 highest and Traffic and ECL lowest by an order of magnitude. Scale energy shift keeps almost the same ranking in both modes, and the conclusion of §5.6 holds under either, that the size of the reshaping does not track whether learning the filter pays.

\newpage
\clearpage
\section{Detailed pseudocode}\label{app:detail}
Algorithm~\ref{alg:wedp_full} gives the full per-step pseudocode. The Fixed, Learned and Dynamic modes differ only as noted in its caption.

\begin{algorithm}[h!]
\caption{Wavelet-energy-driven dynamic patching. \textsc{Fixed}: $(h,g)$ frozen, patches $\{P_i,\sigma_i\}$ computed once before training and cached. \textsc{Learned}: $(h,g)$ trained, patches cached as in Fixed. \textsc{Dynamic} (shown): $(h,g)$ trained, patches recomputed every step. With cached patches the soft weights are not computed, so $\alpha$ has no effect.}
\label{alg:wedp_full}
\begin{algorithmic}[1]
\Require Per-channel filters $\{h^{(c)},g^{(c)}\}_{c=1}^{C}$ initialised to db8 (QMF tie dropped), levels $J$ (bands $S{=}J{+}1$), base-window width $b$, model width $d$, horizon $H$, straight-through strength $\alpha$, temperature $\tau$, maps $W_{\mathrm{in}},W_p,W_s$, tables $\mathrm{PosEmb}$ and $\mathrm{ScaleEmb}$, encoder and forecast head.
\Statex
\While{not converged}
  \State Sample $(X,y)$ with $X\in\mathbb{R}^{C\times L}$ and process each channel $x\in\mathbb{R}^{L}$ independently
  \State $x\gets\mathrm{RevIN}(x)$
  \Step{Step 1: Decomposition and per-scale reconstruction (\S3.2--3.3)}
  \State $(c_{\mathrm a},c_0,\dots,c_{J-1})\gets\mathrm{DWT}_{h,g}(x)$ \Comment{$a_{j+1}=\downarrow_2\!\big(\mathcal F^{-1}(\mathcal F(a_j)\odot{\mathcal F(h)})\big)$, $d_{j+1}$ likewise with $g$}
  \State $r_s\gets\mathrm{IDWT}(0,\dots,c_s,\dots,0)$ for $s{=}0,\dots,J{-}1$, \ $r_{\mathrm a}\gets\mathrm{IDWT}(c_{\mathrm a},0,\dots,0)$ \Comment{each band at length $L$}
  \Step{Step 2: Scale-energy assignment (\S3.4)}
  \For{each base window $w=0,\dots,N{-}1$}\Comment{$w$ spans $[wb,(w{+}1)b)$, $N=\lfloor L/b\rfloor$}
      \State $E_w[s]\gets\sum_{t=wb}^{(w+1)b-1} r_s(t)^2$ \Comment{energy of detail band $s$ in window $w$}
      \State $m_w\gets\arg\max_{s\in\{0,\dots,J-1\}} E_w[s]$ \Comment{hard: dominant detail band}
      \State $\tilde w_w[s]\gets\mathrm{softmax}_s\big(E_w[s]/\tau\big)$ \Comment{soft: differentiable in $(h,g)$}
  \EndFor
  \Step{Step 3: Variable-length patch construction (\S3.4)}
  \State $i\gets 0$, \ $\mathrm{st}\gets 0$ \Comment{open the first patch at window $0$}
  \For{$w=1$ \textbf{to} $N{-}1$}
      \If{$m_w \neq m_{w-1}$}\Comment{dominant band changed, so close the run}
          \State $P_i\gets[\mathrm{st}\cdot b,\ w\cdot b)$, \ $\sigma_i\gets m_{w-1}$
          \State $i\gets i+1$, \ $\mathrm{st}\gets w$
      \EndIf
  \EndFor
  \State $P_i\gets[\mathrm{st}\cdot b,\ L)$, \ $\sigma_i\gets m_{N-1}$, \ $M\gets i+1$ \Comment{the final patch absorbs any remainder of $L$}
  \Step{Step 4: Tokenisation with straight-through band embedding (\S3.5)}
  \State $v_t\gets W_{\mathrm{in}}\,x_t$
  \State $e_i^{\mathrm{hard}}\gets\mathrm{ScaleEmb}[\sigma_i]$, \ $e_i^{\mathrm{soft}}\gets\big(\tfrac1{|W_i|}\sum_{w\in W_i}\tilde w_w\big)\,\mathrm{ScaleEmb}[0{:}J]$
  \State $e_i\gets e_i^{\mathrm{hard}}+\alpha\big(e_i^{\mathrm{soft}}-\mathrm{stopgrad}(e_i^{\mathrm{soft}})\big)$ \Comment{in Dynamic, the only path from $\mathcal L$ to $(h,g)$ through the patching}
  \State $p_i\gets W_p\big[\max;\operatorname{mean};\min\big]_{t\in P_i} v_t + \mathrm{PosEmb}[i] + e_i$ \Comment{$M$ patch tokens $P$}
  \State $u_s\gets W_s\,r_s$ for $s\in\{\mathrm a,0,\dots,J{-}1\}$ \Comment{$S$ scale tokens $U$, coarse to fine}
  \Step{Step 5: Attention, forecast and update (\S3.6--3.7)}
  \State $C\gets\mathrm{LayerNorm}\big(U+\mathrm{CrossAttn}(Q{=}U,\,K{=}V{=}P)\big)\in\mathbb{R}^{S\times d}$
  \State $Z\gets\mathrm{SelfAttn}(C)$ \Comment{$L_e$ layers, causal over the coarse-to-fine order}
  \State $\hat y\gets\mathrm{RevIN}^{-1}\big(\mathrm{Head}(Z)\big)$, \ $\mathcal{L}\gets\tfrac1H\lVert\hat y-y\rVert_2^2$
  \State Update encoder, $W_{\mathrm{in}},W_p,W_s$, $\mathrm{PosEmb}$, $\mathrm{ScaleEmb}$, head via $\nabla\mathcal{L}$, and $\{h^{(c)},g^{(c)}\}$ in Learned and Dynamic
\EndWhile
\end{algorithmic}
\end{algorithm}

\newpage
\section{QMF vs unconstrained filters}\label{app:qmf}
The deployed filter drops the quadrature-mirror (QMF) tie. Under QMF the high-pass $g$ is derived from the low-pass $h$, which ties the two filters together and pins the drift ratio $g/h$ to 1. We instead learn $h$ and $g$ independently. Table~\ref{tab:qmf} compares the two at matched settings (per-feature, p96, best filter width). Dropping the tie helps on three of the four ETT datasets, most on the finely sampled ETTm2 ($+4.2\%$), and is roughly neutral on ETTh1. The mechanism is visible in $g/h$. QMF holds it at 1, so the high-pass cannot move relative to the low-pass, whereas non-QMF lets it rise to 1.9 on ETTm2, adapting the high-frequency response exactly where fast content exists. This is the empirical basis for the unconstrained filter adopted in \S3.2.
\begin{table}[h!]\centering
\caption{Relaxing the QMF constraint (Non-QMF), per-feature, p96, best filter width. \%Imp is QMF$\to$Non-QMF (positive = Non-QMF better); $g/h$ is the high-/low-pass drift ratio under Non-QMF (QMF fixes it at 1). Best per row \textbf{bold}.}
\label{tab:qmf}
\begin{tabular*}{\textwidth}{@{\extracolsep{\fill}}lcccc@{}}
\toprule
Dataset & QMF & Non-QMF & QMF$\to$Non-QMF \%Imp & $g/h$ (Non-QMF) \\
\midrule
ETTh1 & \textbf{0.3671} & 0.3691 & $-0.54$ & 1.27 \\
ETTh2 & 0.2943 & \textbf{0.2879} & $+2.19$ & 1.03 \\
ETTm1 & 0.3238 & \textbf{0.3197} & $+1.27$ & 1.25 \\
ETTm2 & 0.1818 & \textbf{0.1741} & $+4.24$ & 1.91 \\
\bottomrule
\end{tabular*}
\end{table}

\section{Vanishing-moment regularisation}\label{app:vmreg}
As a further check that the filter's exact form is not what drives accuracy, we add a regulariser that penalises non-zero low-order moments of $g$, that is, departure from the vanishing-moment property of the initial wavelet and vary its strength $\lambda$ from 0 to 1 (Table~\ref{tab:vmreg}). Forcing the filter to stay a proper wavelet has essentially no effect on test MSE (coefficient of variation below 0.15\% on every dataset). Wavelet purity is therefore decoupled from forecast quality within a wide equivalence class of filters, an independent check that corroborates the mode-ladder finding that learning the filter is only a marginal effect.
\begin{table}[h!]\centering
\caption{Test MSE as the vanishing-moment penalty $\lambda$ is swept (per-feature, p96).}
\label{tab:vmreg}
\begin{tabular*}{\textwidth}{@{\extracolsep{\fill}}lcccccc@{}}
\toprule
Dataset & $\lambda{=}0$ & $10^{-3}$ & $10^{-2}$ & $0.1$ & $1.0$ & CV\% \\
\midrule
ETTh1 & 0.3702 & 0.3702 & 0.3702 & 0.3702 & 0.3704 & 0.02 \\
ETTh2 & 0.2880 & 0.2880 & 0.2881 & 0.2881 & 0.2881 & 0.01 \\
ETTm1 & 0.3251 & 0.3251 & 0.3249 & 0.3244 & 0.3245 & 0.09 \\
ETTm2 & 0.1745 & 0.1745 & 0.1745 & 0.1744 & 0.1744 & 0.03 \\
Weather & 0.1632 & 0.1632 & 0.1633 & 0.1634 & 0.1638 & 0.14 \\
Exchange & 0.0825 & 0.0825 & 0.0825 & 0.0825 & 0.0825 & 0.00 \\
\bottomrule
\end{tabular*}
\end{table}

\end{document}

%% file: math_commands.tex
\usepackage{amsmath,amsfonts,bm}

\def\eqref#1{equation~\ref{#1}}

\def\1{\bm{1}}

\DeclareMathAlphabet{\mathsfit}{\encodingdefault}{\sfdefault}{m}{sl}
\SetMathAlphabet{\mathsfit}{bold}{\encodingdefault}{\sfdefault}{bx}{n}



%% file: litreview_refs.bib
@inproceedings{zeng2023dlinear,
  title={Are Transformers Effective for Time Series Forecasting?},
  author={Zeng, Ailing and Chen, Muxi and Zhang, Lei and Xu, Qiang},
  booktitle={Proceedings of the AAAI Conference on Artificial Intelligence}, year={2023},
  note={arXiv:2205.13504}}

@inproceedings{nie2023patchtst,
  title={A Time Series is Worth 64 Words: Long-term Forecasting with Transformers},
  author={Nie, Yuqi and Nguyen, Nam H. and Sinthong, Phanwadee and Kalagnanam, Jayant},
  booktitle={International Conference on Learning Representations (ICLR)}, year={2023}, note={arXiv:2211.14730}}

@inproceedings{liu2024itransformer,
    title={{iTransformer}: Inverted Transformers Are Effective for Time Series Forecasting},
  author={Liu, Yong and Hu, Tengge and Zhang, Haoran and Wu, Haixu and Wang, Shiyu and Ma, Lintao and Long, Mingsheng},
  booktitle={International Conference on Learning Representations (ICLR)}, year={2024}, note={arXiv:2310.06625}}

@inproceedings{wu2023timesnet,
  title={{TimesNet}: Temporal 2D-Variation Modeling for General Time Series Analysis},
  author={Wu, Haixu and Hu, Tengge and Liu, Yong and Zhou, Hang and Wang, Jianmin and Long, Mingsheng},
  booktitle={International Conference on Learning Representations (ICLR)}, year={2023}, note={arXiv:2210.02186}}

@inproceedings{chen2024pathformer,
  title={{Pathformer}: Multi-scale Transformers with Adaptive Pathways for Time Series Forecasting},
  author={Chen, Peng and Zhang, Yingying and Cheng, Yunyao and Shu, Yang and Wang, Yihang and Wen, Qingsong and Yang, Bin and Guo, Chenjuan},
  booktitle={International Conference on Learning Representations (ICLR)}, year={2024}, note={arXiv:2402.05956}}

@article{abeywickrama2025entrope,
  title={{EntroPE}: Entropy Guided Dynamic Patch Segmentation for Time Series Transformers},
  author={Abeywickrama, Sachith and Eldele, Emadeldeen and Wu, Min and Li, Xiaoli and Yuen, Chau},
  journal={arXiv preprint arXiv:2509.26157}, year={2025}}

@inproceedings{ding2025timemosaic,
  title={{TimeMosaic}: Temporal Heterogeneity Guided Time Series Forecasting via Adaptive Granularity Patch and Segment-wise Decoding},
  author={Ding, Kuiye and Fan, Fanda and Hou, Chunyi and Wang, Zheya and Wang, Lei and Yang, Zhengxin and Zhan, Jianfeng},
  booktitle={Proceedings of the AAAI Conference on Artificial Intelligence}, year={2026}, note={arXiv:2509.19406}}

@inproceedings{huang2024hdmixer,
  title={{HDMixer}: Hierarchical Dependency with Extendable Patch for Multivariate Time Series Forecasting},
  author={Huang, Qihe and Shen, Lei and Zhang, Ruixin and Cheng, Jiahuan and Ding, Shouhong and Zhou, Zhengyang and Wang, Yang},
  booktitle={Proceedings of the AAAI Conference on Artificial Intelligence}, volume={38}, number={11}, pages={12608--12616},
  year={2024}, doi={10.1609/aaai.v38i11.29155}}

@article{pagnoni2024blt,
  title={Byte Latent Transformer: Patches Scale Better Than Tokens},
  author={Pagnoni, Artidoro and others}, journal={arXiv preprint arXiv:2412.09871}, year={2024}}

@article{zucchi2026,
  title={Adaptive Patching Is Harder Than It Looks For Time-Series Forecasting},
  author={Zucchi, Federico and Xie, Yi and Zhang, Chao and Luo, Keyuan and Lampert, Thomas and Li, Ziyue}, journal={arXiv preprint arXiv:2606.04074}, year={2026}}

@inproceedings{wu2021autoformer,
  title={{Autoformer}: Decomposition Transformers with Auto-Correlation for Long-Term Series Forecasting},
  author={Wu, Haixu and Xu, Jiehui and Wang, Jianmin and Long, Mingsheng},
  booktitle={Advances in Neural Information Processing Systems (NeurIPS)}, year={2021}, note={arXiv:2106.13008}}

@inproceedings{zhou2022fedformer,
  title={{FEDformer}: Frequency Enhanced Decomposed Transformer for Long-term Series Forecasting},
  author={Zhou, Tian and Ma, Ziqing and Wen, Qingsong and Wang, Xue and Sun, Liang and Jin, Rong},
  booktitle={International Conference on Machine Learning (ICML)}, year={2022}, note={arXiv:2201.12740}}

@article{xie2025swift,
   title={{SWIFT}: Mapping Sub-series with Wavelet Decomposition Improves Time Series Forecasting},
  author={Xie, Wenxuan and Cao, Fanpu}, journal={arXiv preprint arXiv:2501.16178}, year={2025}}

@article{wang2025wavetuner,
  title={{WaveTuner}: Comprehensive Wavelet Subband Tuning for Time Series Forecasting},
  author={Wang, Yubo and He, Hui and Niu, Chaoxi and Niu, Zhendong}, journal={arXiv preprint arXiv:2511.18846}, year={2025}}

@inproceedings{murad2025wpmixer,
   title={{WPMixer}: Efficient Multi-Resolution Mixing for Long-Term Time Series Forecasting},
  author={Murad, Md Mahmuddun Nabi and Aktukmak, Mehmet and Yilmaz, Yasin},
  booktitle={Proceedings of the AAAI Conference on Artificial Intelligence}, year={2025}, note={arXiv:2412.17176}}

@inproceedings{wang2018mwdn,
  title={Multilevel Wavelet Decomposition Network for Interpretable Time Series Analysis},
  author={Wang, Jingyuan and Wang, Ze and Li, Jianfeng and Wu, Junjie},
  booktitle={Proceedings of the 24th ACM SIGKDD International Conference on Knowledge Discovery \& Data Mining (KDD)},
  year={2018}, note={arXiv:1806.08946}}

@article{yu2024adawavenet,
  title={{AdaWaveNet}: Adaptive Wavelet Network for Time Series Analysis},
  author={Yu, Han and Guo, Peikun and Sano, Akane}, journal={Transactions on Machine Learning Research (TMLR)},
  year={2024}, note={arXiv:2405.11124}}

@article{michau2022fldwt,
  title={Fully Learnable Deep Wavelet Transform for Unsupervised Monitoring of High-Frequency Time Series},
  author={Michau, Gabriel and Frusque, Ga{\"e}tan and Fink, Olga},
  journal={Proceedings of the National Academy of Sciences (PNAS)}, volume={119}, number={8}, year={2022},
  doi={10.1073/pnas.2106598119}}

@inproceedings{shu2026sonnet,
   title={{Sonnet}: Spectral Operator Neural Network for Multivariable Time Series Forecasting},
  author={Shu, Yuxuan and Lampos, Vasileios}, booktitle={Proceedings of the AAAI Conference on Artificial Intelligence}, year={2026},
  note={arXiv:2505.15312}}

@inproceedings{zhuang2024wavspa,
   title={{WavSpA}: Wavelet Space Attention for Boosting Transformers' Long Sequence Learning Ability},
  author={Zhuang, Yufan and others}, booktitle={Proceedings of Machine Learning Research (PMLR v243)}, year={2024},
  note={arXiv:2210.01989}}

@article{videau2025bytes,
  title={From Bytes to Ideas: Language Modeling with Autoregressive {U-Nets}},
  author={Videau, Mathurin and Youbi Idrissi, Badr and Leite, Alessandro and Schoenauer, Marc and Teytaud, Olivier and Lopez-Paz, David},
  journal={arXiv preprint arXiv:2506.14761}, year={2025}}

@inproceedings{kim2022revin,
  title={Reversible Instance Normalization for Accurate Time-Series Forecasting against Distribution Shift},
  author={Kim, Taesung and Kim, Jinhee and Tae, Yunwon and Park, Cheonbok and Choi, Jang-Ho and Choo, Jaegul},
  booktitle={International Conference on Learning Representations (ICLR)}, year={2022}}

@inproceedings{wang2024timemixer,
    title={{TimeMixer}: Decomposable Multiscale Mixing for Time Series Forecasting},
  author={Wang, Shiyu and Wu, Haixu and Shi, Xiaoming and Hu, Tengge and Luo, Huakun and Ma, Lintao and Zhang, James Y. and Zhou, Jun},
  booktitle={International Conference on Learning Representations (ICLR)}, year={2024}, note={arXiv:2405.14616}}

@inproceedings{wang2024timexer, title={{TimeXer}: Empowering Transformers for Time Series Forecasting with Exogenous Variables}, author={Wang, Yuxuan and Wu, Haixu and Dong, Jiaxiang and Liu, Yong and Long, Mingsheng and Wang, Jianmin}, booktitle={Advances in Neural Information Processing Systems (NeurIPS)}, year={2024}, note={arXiv:2402.19072}}

@inproceedings{vaswani2017transformer,
  title={Attention Is All You Need},
  author={Vaswani, Ashish and Shazeer, Noam and Parmar, Niki and Uszkoreit, Jakob and Jones, Llion and Gomez, Aidan N. and Kaiser, {\L}ukasz and Polosukhin, Illia},
  booktitle={Advances in Neural Information Processing Systems (NeurIPS)}, year={2017}, note={arXiv:1706.03762}}
